\documentclass[letterpaper]{article} % DO NOT CHANGE THIS
\usepackage{aaai25}  % DO NOT CHANGE THIS
\usepackage{times}  % DO NOT CHANGE THIS
\usepackage{helvet}  % DO NOT CHANGE THIS
\usepackage{courier}  % DO NOT CHANGE THIS
\usepackage[hyphens]{url}  % DO NOT CHANGE THIS
\usepackage{graphicx} % DO NOT CHANGE THIS
\usepackage{natbib}  % DO NOT CHANGE THIS AND DO NOT ADD ANY OPTIONS TO IT
\usepackage{caption} % DO NOT CHANGE THIS AND DO NOT ADD ANY OPTIONS TO IT
\usepackage{algorithm}
\usepackage{algorithmic}

\usepackage{booktabs}
\usepackage{subcaption}
\usepackage{amsmath}
\usepackage{amssymb}

\newcommand{\etal}{\emph{et al}.}
\newcommand{\ie}{\emph{i.e}.}
\newcommand{\eg}{\emph{e.g}.}
\newcommand{\cref}[1]{Fig.~\ref{#1}}
\newcommand{\fref}[1]{Fig.~\ref{#1}}

\newcommand{\jychecked}[1]{#1}
\newcommand{\heichecked}[1]{#1}
\usepackage{multirow}
\usepackage{makecell}

\usepackage{newfloat}
\usepackage{listings}
\DeclareCaptionStyle{ruled}{labelfont=normalfont,labelsep=colon,strut=off} % DO NOT CHANGE THIS
\floatstyle{ruled}
\newfloat{listing}{tb}{lst}{}
\floatname{listing}{Listing}
\title{Leveraging RGB-D Data with Cross-Modal Context Mining for Glass Surface Detection}
\author{
    Jiaying Lin\equalcontrib \ \ \ \ \ 
    Yuen-Hei Yeung\equalcontrib \ \ \ \ \  
    Shuquan Ye\thanks{Shuquan Ye and Rynson Lau are joint corresponding authors.} \ \ \ \ \ 
    Rynson W.H. Lau\footnotemark[2]
}
\affiliations{
City University of Hong Kong
}

\usepackage{bibentry}
\begin{document}

\maketitle

\begin{abstract}
Glass surfaces are becoming increasingly ubiquitous as modern buildings tend to use a lot of glass panels. This, however, poses substantial challenges to the operations of autonomous systems such as robots, self-driving cars, and drones, as these glass panels can become transparent obstacles to navigation. Existing works attempt to exploit various cues, including glass boundary context or reflections, as priors. However, they are all based on input RGB images. We observe that the transmission of 3D depth sensor light through glass surfaces often produces blank regions in the depth maps, which can offer additional insights to complement the RGB image features for glass surface detection. 
In this work, we first propose a large-scale RGB-D glass surface detection dataset, \textit{RGB-D GSD}, for rigorous experiments and future research. It contains 3,009 images, paired with precise annotations, offering a wide range of real-world RGB-D glass surface categories. We then propose a novel glass surface detection framework combining RGB and depth information, with two novel modules: a cross-modal context mining (CCM) module to adaptively learn individual and mutual context features from RGB and depth information, and a depth-missing aware attention (DAA) module to explicitly exploit spatial locations where missing depths occur to help detect the presence of glass surfaces.
Experimental results show that our proposed model outperforms state-of-the-art methods. 
\end{abstract}

% In this paper, we propose a novel framework for glass surface detection by incorporating RGB-D information, with two novel modules: a cross-modal context mining (CCM) module to adaptively learn individual and mutual context features from RGB and depth information, and a depth-missing aware attention (DAA) module to explicitly exploit spatial locations where missing depths occur to help detect the presence of glass surfaces. Besides, we propose a large-scale RGB-D glass surface detection dataset, called \textit{RGB-D GSD}, which comprises 3,009 real-world RGB-D glass surface images with precise annotations. 

\begin{links}
    \link{Code, Dataset and Extended version}{https://jiaying.link/AAAI25-RGBDGlass/}
    % \link{Extended version}{https://arxiv.org/pdf/xxxx.xxxxx}
\end{links}

\section{Introduction}
\label{sec:intro}

\begin{figure}[t] 
    \centering
    \renewcommand{\tabcolsep}{1pt}
    \newcommand{\imgpath}{teaser_selected/0images/}
    \newcommand{\depthpath}{teaser_selected/depths_colored/}
    \newcommand{\linpath}{teaser_selected/CVPR2021-Lin/}
    \newcommand{\eblnetpath}{teaser_selected/ICCV2021-EBLNet/}
    \newcommand{\rfenetpath}{teaser_selected/IJCAI2023-RFENet/}
    \newcommand{\gtpath}{teaser_selected/masks/}
    \newcommand{\ourpath}{teaser_selected/ours_best_130/}
    \newcommand{\imgwidth}{0.073\textwidth}
    \begin{center}
    \begin{tabular}{@{}cccccccc@{}}
    \gdef\imgid{2f59933fa37143049c27c412c3133c55_2_2}
    \includegraphics[width=\imgwidth]{\imgpath \imgid .jpg} &
    \includegraphics[width=\imgwidth]{\depthpath \imgid .jpg} &
    \includegraphics[width=\imgwidth]{\linpath \imgid .png} &
    \includegraphics[width=\imgwidth]{\eblnetpath \imgid .png} &
    \includegraphics[width=\imgwidth]{\ourpath \imgid .png} &
    \includegraphics[width=\imgwidth]{\gtpath \imgid .png}
    \vspace{-0.5mm}
    \\

    % \gdef\imgid{80f652ffcdfa47ada388a00233624799_1_2}
    % \includegraphics[width=\imgwidth]{\imgpath \imgid .png} &
    % \includegraphics[width=\imgwidth]{\depthpath \imgid .png} &
    % \includegraphics[width=\imgwidth]{\linpath \imgid .png} &
    % \includegraphics[width=\imgwidth]{\eblnetpath \imgid .png} &
    % \includegraphics[width=\imgwidth]{\ourpath \imgid .png} &
    % \includegraphics[width=\imgwidth]{\gtpath \imgid .png}
    % \\  
    
    \gdef\imgid{00002142}
    \includegraphics[width=\imgwidth]{\imgpath \imgid .png} &
    \includegraphics[width=\imgwidth]{\depthpath \imgid .png} &
    \includegraphics[width=\imgwidth]{\linpath \imgid .png} &
    \includegraphics[width=\imgwidth]{\eblnetpath \imgid .png} &
    \includegraphics[width=\imgwidth]{\ourpath \imgid .png} &
    \includegraphics[width=\imgwidth]{\gtpath \imgid .png}
    \vspace{-0.5mm}
    \\  
        
   \gdef\imgid{cc29839c645d4d0387339c57dffda76c_1_3}
    \includegraphics[width=\imgwidth]{\imgpath \imgid .jpg} &
    \includegraphics[width=\imgwidth]{\depthpath \imgid .png} &
    \includegraphics[width=\imgwidth]{\linpath \imgid .png} &
    \includegraphics[width=\imgwidth]{\eblnetpath \imgid .png} &
    \includegraphics[width=\imgwidth]{\ourpath \imgid .png} &
    \includegraphics[width=\imgwidth]{\gtpath \imgid .png}
    \vspace{-1mm}
    \\ 

    \makebox[\imgwidth]{\scriptsize Image} &
    \makebox[\imgwidth]{\scriptsize Depth} &
    \makebox[\imgwidth]{\scriptsize GSDNet} &
    \makebox[\imgwidth]{\scriptsize EBLNet} &
    \makebox[\imgwidth]{\scriptsize Ours} &
    \makebox[\imgwidth]{\scriptsize GT}
    \\
    \end{tabular}
    \end{center}
    \vspace{-5mm}
    \caption{Advantages of detecting glass surfaces with RGB-D images. These examples show that the depth map can provide a strong cue for glass surface detection. State-of-the-art methods, GSDNet~\cite{GSD:2021} and EBLNet~\cite{He_2021_ICCV}, relying only on input RGB images are not able to correctly separate the glass surfaces from the background. Through learning the cross-modal contexts and the correlation between depth-missing regions and glass surface regions, our proposed model can detect the glass surfaces accurately in all three challenging scenes. Note that red regions in the depth images represent missing depths.
    % The second, third and fourth rows show three very challenging scenes with glass surfaces, where the glass surfaces may not be easily identified even for humans. State-of-the-art methods either over- or under-detect the glass surfaces. Through learning the cross-modal contexts and the correlation between depth-missing regions and glass surface regions, our proposed model can detect the glass surfaces accurately in all three scenes. Note that red regions in the depth images represent missing depths.
    }
    \label{fig:teaser}
    \vspace{-4mm}
\end{figure}

Although most objects that we see in our daily life exhibit distinctive visual characteristics, \jychecked{glass surfaces, however, typically do not possess any distinctive visual properties. Their contents are essentially represented by the contents behind them, making it difficult for existing computer vision systems to detect them. However, modern architects and designers tend to use glass panels frequently in their designs, which makes them prevalent both indoors and outdoors, \eg, glass doors/windows, shop windows, and glass walls. Hence, detecting these glass panels accurately is crucial for the safe operation of autonomous machines such as robots and drones.}
% Early studies on glass surface detection try to leverage priors, \eg, visual cues~\cite{1467548}. Later on, rich sensor data is incorporated, including laser sensor \cite{4543667, WANG201797} and LiDAR sensor \cite{DBLP:journals/corr/abs-1909-12483, Tibebu_2021}, for computations using point clouds and object mapping. 
% A recent method~\cite{10.1145/3197517.3201319} uses fiducial markers to help identify the presence of mirror and glass surfaces, but requires special hardware for tagging. 

The potential of deep learning models for glass surface detection (GSD) was first demonstrated by \cite{Mei_2020_CVPR} and GSDNet\cite{GSD:2021}. 
EBLNet~\cite{He_2021_ICCV} was proposed for GSD through enhanced boundary learning on glass surfaces. However, since EBLNet is an RGB-based method and relies heavily on boundary learning, it may fail if the boundaries of glass surfaces are ambiguous or there are some distracting glass-like regions in the input image. 
Fig.~\ref{fig:teaser} shows three challenging scenarios where state-of-the-art RGB-based glass surface detection methods fail. 
In the first row, both GSDNet and EBLNet are confused by the similarity between glass regions and tiles. As a result, some tiles are mis-recognized as glass regions. 
The other rows show challenging scenes where some glass surfaces are difficult even for humans to identify. GSDNet and EBLNet either over- or under-detect them. 
% In particular, in the second row, the appearance of the room entrance in the middle is very similar to the glass wall of the room on the right hand side. 
%Both state-of-the-art methods partially mis-detect it as glass regions. 

In this paper, we aim to address \jychecked{the limitation of RGB-based glass surface detection methods, which heavily rely on boundary learning, causing them to fail if the glass boundaries are ambiguous or there are glass-like regions in the image}. Here, we propose to address this problem from a depth-aware perspective.
Our key observation is that compared with the \textit{correct} depth of a glass surface, the depth \textit{captured} by a depth sensor typically has two properties, due to the transmission and possibly reflection of the glass surface: 1) the captured depth is noisier, and 2) missing depths frequently appear around the surface.
These two properties indicate that glass surfaces have different contextual characteristic representations in the depth map compared with those of the RGB image. 
Our insight from this observation is that cross-modal context and spatial information of missing depths can provide strong cues for glass surface detection.
This motivates us to design a novel depth-aware glass surface detection method with two new modules: a cross-modal context mining (CCM) module to adaptively fuse multi-modal data by utilizing depth scans coupled with RGB data, and a depth-missing aware attention (DAA) module to explicitly learn the correlation between missing depths and glass regions.

In order to train our model, we need to have a RGB-D dataset for glass surfaces. Although there is an RGB-D dataset~\cite{seib2017friend} available for transparent object detection, it is not suitable for our purpose for two reasons.
First, this dataset targets localizing \jychecked{small} transparent objects with limited patterns of shapes and object types, while we focus on a more challenging task of detecting glass surfaces without well-defined shapes. Second, it contains only 440 RGB-D images \jychecked{captured from a single scene} with \jychecked{four object types} (\ie, beer mugs, water glass, white beer glass, and wine glass), limiting real-world applicability. To address these limitations, we have constructed a new large-scale RGB-D glass surface dataset from diverse scenes with glass surfaces. It contains 3,009 RGB-D images with glass surfaces and corresponding annotated masks. A dataset of this scale allows robust model training and evaluation.
Extensive experiments are conducted to evaluate our method, in comparison with the state-of-the-art methods from relevant tasks. The proposed model outperforms existing methods on our proposed dataset. 
% We will release our RGB-D dataset and codes to the public to facilitate further research.
% Our RGB-D dataset and codes can be found in the project page: https://jiaying.link/AAAI25-RGBDGlass/

Our contributions can be summarised as follows:
\begin{enumerate}
    \item We propose a framework for glass surface detection by  incorporating the depth information, with two novel modules:
    a CCM module to jointly learn the RGB context features and depth context features for comprehensive RGB-D context modeling, and a DAA module to explicitly exploit the locations where the depth information is missing in the depth maps for glass surface detection.
    \item We have constructed a large-scale glass surface dataset of 3,009 images from diverse real-world scenes, with corresponding depth maps and ground truth labels.
    \item Extensive experiments demonstrate the superior performances of our proposed method over SOTA methods.
\end{enumerate}

\section{Related work}
\label{sec:related_work}

\noindent\textbf{Glass Surface Detection.}
% Early methods on glass surface detection are mainly based on hand-craft features. As LiDAR is not able to capture signals from specular surfaces, \eg, glass surfaces, at most angles, Foster \etal~\cite{6630875} propose refinement on an occupancy-grid algorithm to distill glass object surfaces. In view of the inability of LiDAR sensors to properly trace transparent surfaces, ultrasonic scans are incorporated in \cite{8584213, 8664805}, along with fusion on depth scans \cite{ 8584213, Huang2018GlassDA}.\jychecked{These heuristic approaches require various kinds of sensors and hand-craft features, and do not take advantages of the deep-learning technologies, resulting in unsatisfactory performances.}
\cite{Mei_2020_CVPR} propose the first deep-learning model for detecting glass surfaces, given just an RGB image. This method does not require any special hardware. 
As this model relies heavily on contextual contrast learning, it likely fails in complex scenes with insufficient contexts. \cite{GSD:2021} extend contextual contrast learning with modules for detecting glass boundaries and reflections.
However, if the glass surfaces lack reflections or have ambiguous boundaries, the model may not be able to detect the glass surfaces correctly. Similarly, \cite{He_2021_ICCV} propose EBLNet based on learning the glass boundaries. 
Later, RFENet is proposed~\cite{RFENet}  by further exploiting edges in a reciprocal way.
More recently, Liu~\etal propose a new dataset~\cite{liu2024multi} for video glass surface detection.
However, all these deep-learning based methods only consider the RGB information for glass surface detection, and can easily be confused by the distracting glass-like regions that look like glass surfaces, \eg, a door frame. In contrast, we propose in this paper a multi-modal glass surface detection method, based on using RGB-D images as input. Our experimental results show that the proposed method is more robust and accurate.

\vspace{0.5mm}

\noindent\textbf{Transparent Object Detection.}
\jychecked{Compared with glass surface detection, research on transparent object detection (TOD) focuses on detecting small transparent objects, such as wine glass and glass balls, which typically have classic shapes or boundary properties.}
% Adelson and An~\cite{Adelson90ordinalcharacteristics} study the optical characteristics of transparency. Murase~\cite{139539} studies image pattern distortion of moving an object behind a transparent layer by leveraging optical flow. Hata \etal~\cite{547652} and Wang \etal~\cite{inproceedings} try to extract the shapes of transparent objects, while Fritz \etal~\cite{NIPS2009_e46de7e1} \jychecked{study the local patch structure of transparent objects.} 
\jychecked{Some recent TOD methods integrate more accurate measurement data, including data from light-field sensors~\cite{Xu_2015_ICCV} and from infra-red sensors~\cite{2019, 10.1117/12.2266255}}.
Some latest methods use auxiliary information, including polarization~\cite{Kalra_2020_CVPR} and specular reflection~\cite{rs13030455}, to assist the detection.
Other methods use explicit boundary maps \cite{xie2020segmenting} for training, or a transformer architecture \cite{xie2021segmenting, DBLP:journals/corr/abs-2108-09174} for improved performances. However, while the former induces additional data preparation and computation costs, the latter further abstracts the object representations with a feature embedding dictionary and has a constrained set of prototype categories. 
Recently, some datasets~\cite{sun2023trosd,chen2022clearpose,ramirez2022open,ramirez2023booster} have been proposed for this task, but they are still limited to specific objects.
In general, transparent objects are mostly small and with specific shapes (i.e., bottle, wine glass), and tend to have different lighting properties along the boundary. 
% Hence, most TOD methods exploit the boundary cue or some auxiliary information to achieve good performances. 
However, glass surfaces generally do not possess these properties, making these TOD methods not suitable for detecting glass surfaces.

\section{RGB-D GSD Dataset}
% There are many existing datasets \cite{dai2017scannet, Silberman:ECCV12, McCormac:etal:ICCV2017, Janoch:EECS-2012-85, 2017arXiv170201105A, 7298655, Matterport3D} that incorporate depth information, \jychecked{for other tasks such as semantic segmentation and salient object detection.
% }
Existing datasets~\cite{Mei_2020_CVPR, GSD:2021} for glass surface detection do not include depth information, even though it is useful for this task.
\jychecked{In order to train our model and to encourage further research, we propose an RGB-D Glass Surface Detection (RGB-D GSD) dataset.}

\vspace{0.5mm}

\begin{figure}[t]
  \centering \includegraphics[width=0.43\textwidth]{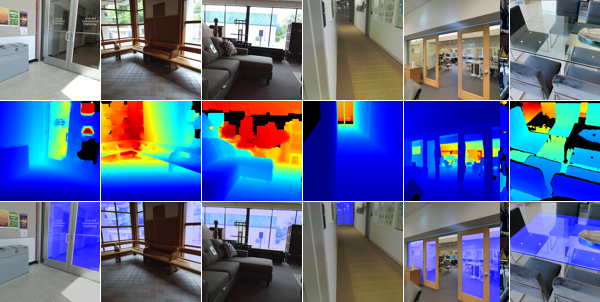}
  \vspace{-2mm}
  \caption{Examples from our RGB-D GSD dataset. Top, middle and bottom rows show RGB images, depth maps, GT glass surface masks overlaid on the images, respectively.}
  \label{fig:glassrgbd}
\end{figure}

\noindent\textbf{Dataset construction.} This dataset contains a total of 3,009 images, where 2,400 for training and 609 for testing. It is an attentively curated ensemble of \jychecked{three existing datasets \jychecked{originally developed for scene understanding}}, including SUN RGB-D \cite{2017arXiv170201105A}, 2D-3D-Semantics \cite{7298655} and Matterport3D \cite{Matterport3D}. Thus, our dataset offers a wide range of glass surface categories, such as window, door, wall, table, cabinet and guardrail. 
For \jychecked{these datasets to be consistent, we normalize all depth images to the range [0, $2^{16} - 1$], and set all invalid depth values to be the minimum value (\ie, 0). In addition, as these datasets do not aim at labeling glass surfaces, they contain inaccurate glass surface labels. \jychecked{For example, the objects inside a glass surface instead of the glass surface itself were labeled;  handles, frames or blinds were labeled as part of the glass surface.}
Hence, we \textbf{manually relabel} the glass surfaces in these images.
}
Figure \ref{fig:glassrgbd} shows some examples from our RGB-D GSD dataset.

\vspace{0.5mm}

\noindent\textbf{Dataset analysis.} We analyze the datasets with the following statistical metrics:
% \begin{figure}
%     \centering
%     \begin{tabular}{cc}
%     \multirow{2}{*}{\includegraphics[width=0.2\textwidth]{charts/area_overlap.png}} &
%     \includegraphics[width=0.1\textwidth]{charts/color_contrast_redraw.png} \\
%     & \includegraphics[width=0.1\textwidth]{charts/area_ratio_redraw.png} \\
%     \makebox[0.2\textwidth]{ (a) Location distribution} &
%     \makebox[0.1\textwidth]{ (b) Color contrast} \\
%     & \makebox[0.1\textwidth]{ (c) Area ratio} \\
%     \end{tabular}
%     \caption{Statistics of our proposed dataset.}\label{fig:ABCD}
% \end{figure}

\begin{figure}[t]
\vspace{-3mm}
    \centering
    \newcommand{\vspacesize}{-2}
    \newcommand{\hspacesize}{-3}
    \includegraphics[width=0.2\textwidth]{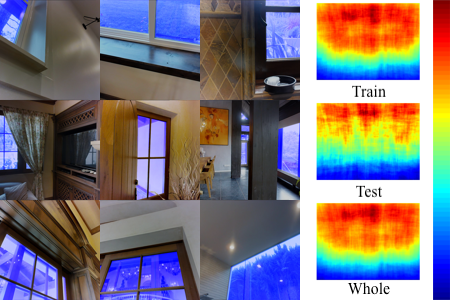}
    \includegraphics[width=0.12\textwidth]{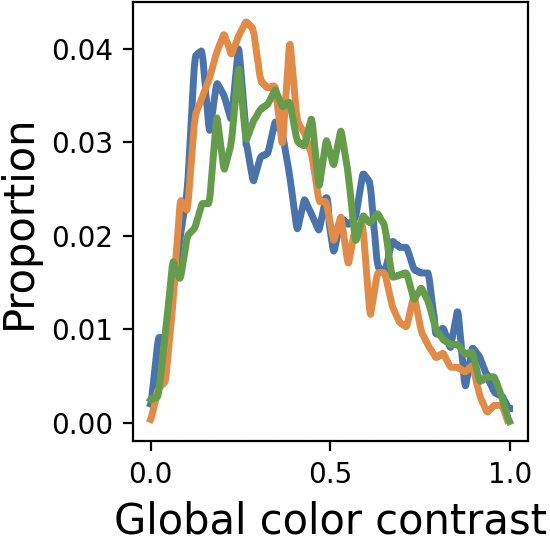}
    \includegraphics[width=0.12\textwidth]{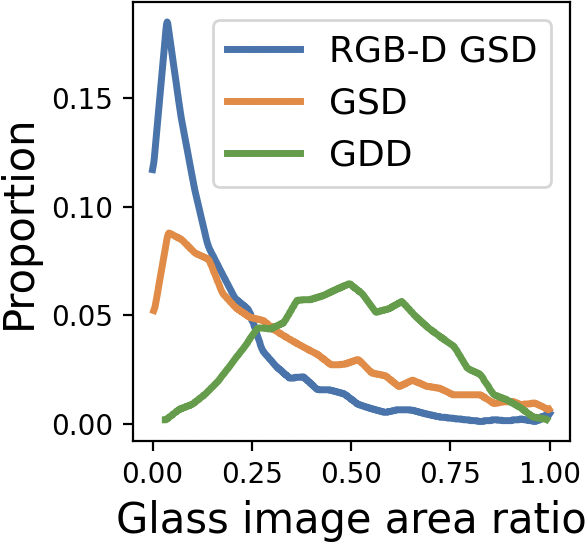}\\
    \hspace{-2mm}\makebox[0.25\textwidth]{\footnotesize(a) Location distribution}
    \makebox[0.2\textwidth]{\footnotesize (b) Color contrast and area ratio}
    \\
    \vspace{-2mm}
    \caption{Statistics of our proposed dataset.}
    \label{fig:ABCD}
    \vspace{-3mm}
\end{figure}

\begin{itemize}
\item \textbf{Glass Location.} The glass location distribution is the average of all glass surface regions in the dataset. The maps in Figure \ref{fig:ABCD}(a) show that glass surfaces mainly concentrate at the top region, which is consistent in the training and testing splits. This also avoids the ``center bias'' problem due to natural observation tendency.

\item \jychecked{\textbf{Color Contrast.} The color contrast between glass and non-glass regions should ideally be low. Otherwise, salient color features can skew the glass surface detection task. We measure the color contrast by computing the $\chi^2$ distance of the RGB histograms between glass and non-glass regions. Figure \ref{fig:ABCD}(b) compares the color contrast among GDD \cite{Mei_2020_CVPR}, GSD \cite{GSD:2021} and our RGB-D GSD. In general, the contrast values of our RGB-D GSD images concentrate in the lower quartile (0 $<$ contrast $<$ 0.4), \jychecked{which is similar to the other two datasets}.}

\item \textbf{Area Ratio.} This metric measures the glass region size relative to the image size. This illustrates the level of semantic context that the images provide. In other words, a smaller glass region leaves more room for the surrounding environment to offer additional hints. As mentioned in GSD, GDD contains primarily close-up shots, which limits the amount of contextual information. In addition, in real-life scenario, an autonomous system is expected to be able to detect objects and perform scene understanding tasks as early as possible. Therefore, data of low area ratio is much more meaningful and beneficial for model training. In Figure \ref{fig:ABCD}(b), RGB-D GSD has more images with small glass areas than the other datasets.

\item \textbf{Missing Depth.} Missing depths strongly correlate with the presence of glass surfaces, as glass often causes erratic depth signals. To validate this, we calculated the ratio $\frac{\mathsf{M} \cap \mathsf{G}}{\mathsf{M}}$,
where \(\mathsf{M}\) denotes the set of missing depths and \(\mathsf{G}\) denotes the set of ground truth glass regions.
This ratio represents the alignment of missing depths with ground truth glass regions. The distribution is as follows: 0.0-0.2: 15.47\%, 0.2-0.4: 14.28\%, 0.4-0.6: 14.49\%, 0.6-0.8: 18.74\%, 0.8-1.0: 37.02\%. These results highlight that missing depths, especially at higher ratios, strongly overlap with ground truth, confirming their association with glass surfaces.
\end{itemize}

\section{Method}

\begin{figure*}
    \centering
    \includegraphics[width=0.85\textwidth]{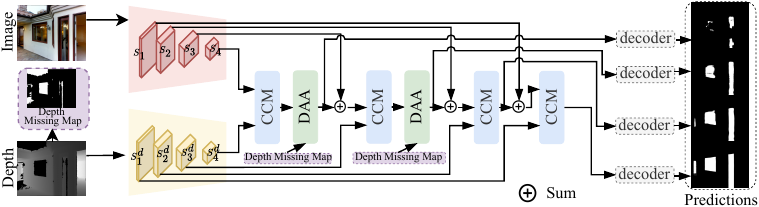}
    \vspace{-4mm}
    \caption{\jychecked{The pipeline of our proposed framework. 
    % Given an input image and a corresponding depth map, our method first extracts multi-scale image features and depth features with a RGB backbone network
    % % ~\cite{xie2017aggregated} 
    % and a depth backbone network
    % % shown in Table~\ref{table:depth_backbone}
    % . It  then uses a novel CCM module (blue blocks) in each stage to learn cross-modal contextual features. For cross-modal contextual features in \textit{stage4} and \textit{stage3}, it further uses a DAA module (green blocks) to enhance them by exploiting the spatial information of the missing depths. Finally, it uses a decoder to extract the prediction of glass surfaces at each scale.
    }}
    \label{fig:framework}
\end{figure*}

\fref{fig:framework} shows the architecture of our proposed framework.
% for RGB-D glass surface detection. 
% The proposed framework consists of four major components: the backbone network for the input RGB images (in red), the backbone network for the input depth maps (in yellow), the cross-modal context mining (CCM) modules (in blue), and the depth-missing aware attention (DAA) modules (in green). These components are arranged to enable multi-stage feature learning with bottom-up and top-down information flows.
In the RGB backbone (red), we first feed the input RGB image to the backbone network~\cite{xie2017aggregated} to extract multi-scale RGB backbone features. Specially, the last four stages' outputs, \ie, $\{S_i\}_{i=1}^{4}$
% \textit{conv1}, \textit{conv2}, \textit{conv3}, and \textit{conv4}
is used as our RGB backbone features. 
In the depth backbone (yellow), 
% unlike the RGB images, we adopt a different backbone network to extract depth features from the input depth map, as shown in Table~\ref{table:depth_backbone}.
the network is simpler and lighter than the RGB one due to depth maps' sparser information and efficiency considerations. 
% the depth backbone network that we use to extract depth features from the input depth map, is much simpler and lighter, compared to the RGB one. There are two reasons. First, using a lighter depth backbone network makes our full framework more efficient in both training and test stages. Second, we observe that depth maps contain sparser information. Simply adopting the same network as the RGB image for the depth map may cause a modality gap between the RGB and depth information, which will lead to performance degradation. 
Similarly, we only use the last four stages' outputs, denoted as
$\{S^{d}_i\}_{i=1}^{4}$.
% \textit{depth\_conv1}, \textit{depth\_conv2}, \textit{depth\_conv3}, and \textit{depth\_conv4}.
% After obtaining the RGB backbone features and the depth backbone features, we first feed the \jychecked{uppermost} RGB and depth backbone features (\ie, \textit{conv4} and \textit{depth\_conv4})
We first feed the RGB and depth backbone features (\ie, $S_4$ and $S^{d}_4$) into a CCM module to capture the multi-modal contextual features comprehensively at the \jychecked{latest} stage (\ie, \textit{stage4}). 
\jychecked{We create a binary depth-missing map from the depth map by \jychecked{setting} the invalid depth pixels to 1's and valid pixels to 0's.}
The multi-modal contextual features \jychecked{from CCM} and the resized depth-missing map are then fed into the DAA module to enhance the contextual features by exploiting the \jychecked{spatial location information of missing depths}. Finally, using the enhanced contextual features the decoder produces a coarse binary mask representing \textit{stage4}'s detected glass surfaces. 
\jychecked{In the preceding stage (\ie, \textit{stage3}), the enhanced contextual features from \textit{stage4} are then added to the RGB features (\ie, $S_3$)}, before feeding into CCM, DAA, and decoder.
We repeat this process in \textit{stage2} and \textit{stage1}, but omitting depth missing attention for two reasons.
% First, the depth-missing map that we use indicates spatial locations where depths are missing. 
First, while such spatial information can be easily learned from \jychecked{the upper layers (\ie, later stages)}, it is harder to learn from \jychecked{the lower layers (\ie, earlier stages)}. 
% Thus, \jychecked{we apply the DAA module in the two upper layers to allow our framework to} exploit low- and high-level information effectively. 
% \jychecked{Second, as the spatial size of the features increases from the later stages to the earlier stages, }
\jychecked{Second, adopting additional components to the early stages will heavily decrease the efficiency of the proposed network.}
With this progressive refinement, the earliest stage (\ie, \textit{stage1}) outputs the finest binary mask \jychecked{as the final output}. More details (\eg, the depth backbone) are included in the \textit{supplementary material} due to limited space.

% \jychecked{For the rest of this section, we first discuss how the proposed CCM module exploits multi-modal contextual features to adaptively learn individual and mutual context features in Section~\ref{sec:ccmm}, and how the proposed DAA module takes advantage of the depth-missing information to help detect glass surfaces in Section~\ref{sec:dmaa}. We then describe the loss functions that we use to train the whole framework.}

\begin{figure*}[hbt!]
    \centering
    \vspace{-3mm}
    
    \includegraphics[width=0.9\textwidth]{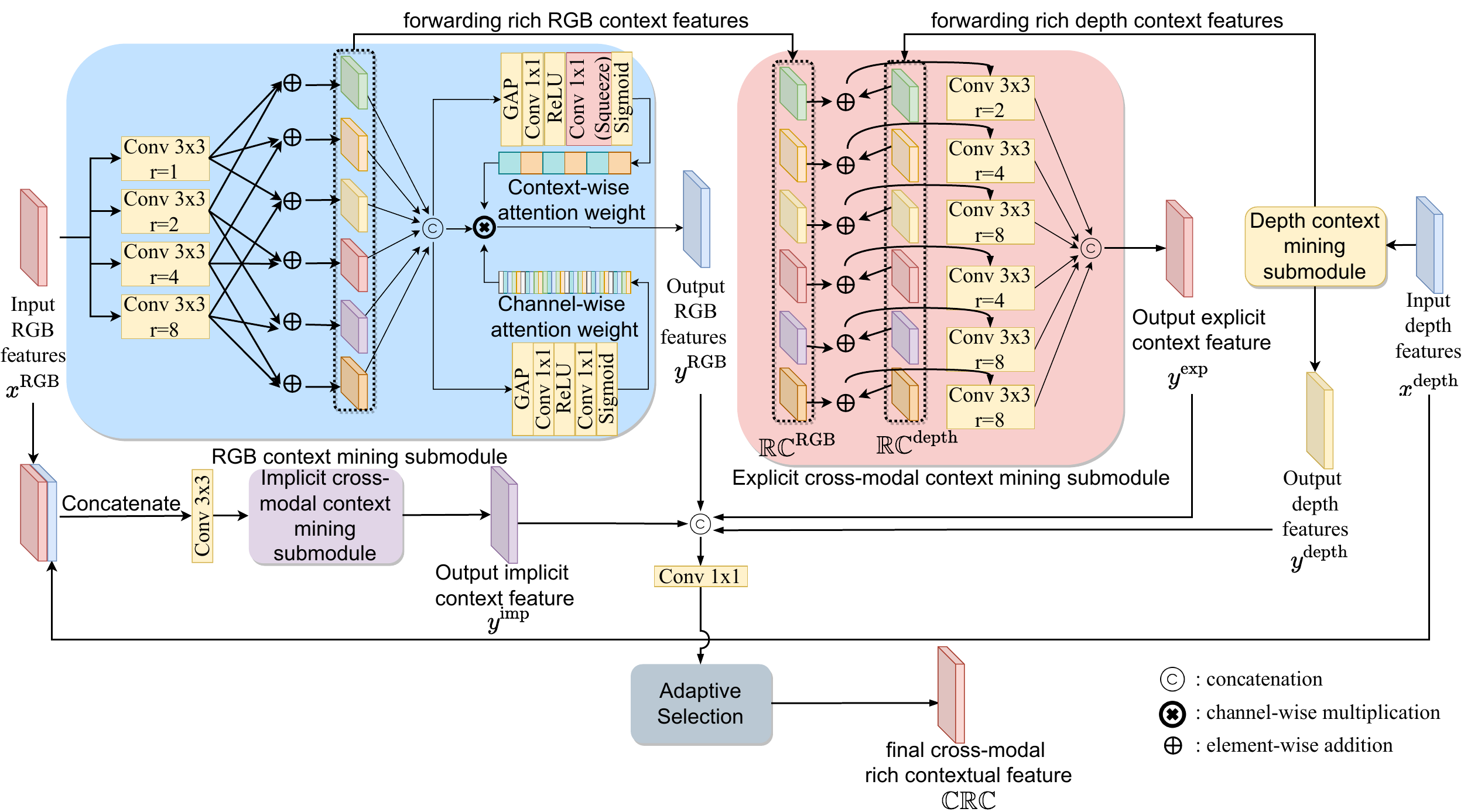}
    \vspace{-3.5mm}
    \caption{Illustration of our proposed cross-modal context mining (CCM) module.
    % \jychecked{It consists of four submodules: a RGB context mining submodule, a depth context mining submodule, an implicit multi-modal context mining submodule, and an explicit multi-modal context mining submodule. The CCM module first takes the RGB and depth features as input and outputs intermediate contextual features with its four submodules. These intermediate features are then concatenated and adaptively selected to generate the final cross-modal rich context features as the output of the CCM module.} 
    }
    \label{fig:ccmm}
\end{figure*}

\subsection{Cross-modal Context Mining (CCM) Module}\label{sec:ccmm}
Previous works~\cite{Mei_2020_CVPR, GSD:2021} 
% on glass surface detection show that contextual information is very useful for the task. However, these works 
\jychecked{focus only on single-modal glass surface detection. To exploit multi-modal data in this problem,}
we design the CCM module to adaptively learn the context features from both RGB and depth information. 
\fref{fig:ccmm} illustrates the structure of the proposed CCM module. It consists of four submodules to model the multi-modal context from different aspects. The outputs of these four submodules are concatenated for adaptive selection to produce the final cross-modal feature.

\vspace{0.5mm}

\noindent\textbf{RGB and Depth Context Mining Submodule.}
Given the input RGB features $x^\text{RGB} \in \mathbb{R}^{C \times H\times W}$, this submodule extracts a series of multi-scale context features $\mathbb{C}_{r}^{\text{RGB}}$ by atrous convolutions with \jychecked{dilation} rate $r \in [1, 2, 4, 8]$. These multi-scale context features are then fused mutually by element-wise addition to form an interim rich representation of contextual information ($\mathbb{RC}^{\text{RGB}}$). After obtaining all permuted pairs of different context scales, these feature pairs are then concatenated to produce the aggregated rich contextual features for the RGB information $\mathbb{ARC}_{\text{RGB}}$. The whole process of this context aggregation operation is:
\begin{equation}
\begin{split}
        \mathbb{RC}_{r_i,r_j}^{\text{RGB}}  &= \mathbb{C}_{r_i}^{\text{RGB}} + \mathbb{C}_{r_j}^{\text{RGB}} (r_i < r_j) , \\
\end{split}
\end{equation}
where $r_i$ and $r_j$ are two different \jychecked{dilation} rates used to produce the multi-scale context features $\mathbb{C}$.

\jychecked{The aggregated rich contextual features for the RGB information $\mathbb{ARC}_{\text{RGB}}$ is then} computed as:
\begin{equation}
        {\mathbb{ARC}^{\text{RGB}} = \text{Concat}(\underbrace{\mathbb{RC}_{1,2}^{\text{RGB}}, \mathbb{RC}_{1,4}^{\text{RGB}}, ...,\mathbb{RC}_{2,8}^{\text{RGB}}, \mathbb{RC}_{4,8}^{\text{RGB}}
        }_{\binom{4}{2}})},
\end{equation}
$\mathbb{ARC}^{\text{RGB}}$ are then forwarded to a channel-wise attention (CNA) and a context-wise attention (CXA). 
% The CNA mechanism that we use in the submodule consists of an average pooling layer and two convolution layers with a ReLU and sigmoid activation, as:
% \begin{equation}
%     CNA(x) = x \times \sigma(\psi_2(ReLU(\psi_1(\mu(x))))) ,
% \end{equation}
% \jychecked{where $\mu$, $ReLU$, $\sigma$ and $\psi$ are the global average pooling (GAP) layer, ReLU, sigmoid function and convolution layers with a $1\times 1$ kernel, respectively. $\psi_1$ and $\psi_2$ are $1\times 1$ convolution layers with different weights.}
% $x$ represents the input features. The output of channel-wise attention has the same number of channels as the input features $x$.
% Similarly, we can obtain the CXA by adjusting the output channels of the convolution layers that we use. 
Unlike CNA, which computes individual weights for different channels in the input features, the attention weights \jychecked{for} the CXA are shared across all channels within the same context by squeezing and broadcasting. We finally \jychecked{multiply the channel-wise attention weights, the context-wise attention weights and the input features together to form the} final output $y^\text{RGB}$. 
% \noindent\textbf{Depth Context Mining Submodule.}
Similar to the RGB context mining submodule, we can obtain the aggregated rich contextual features for the depth information $\mathbb{ARC}^{\text{depth}}$ by the same module design. 
% The difference between the RGB context mining submodule and the depth context mining submodule is that the former submodule takes the RGB backbone features $x^\text{RGB}$ as input, while the later submodule takes the depth backbone features $x^\text{depth}$. The model weights of these two submodules are not shared, so that they can focus on context mining in \jychecked{their own} modalities. 
% We also apply the channel-wise attention and the context-wise attention on $\mathbb{ARC}^{\text{depth}}$ to produce the final output $y^\text{depth}$ of this submodule.

\vspace{0.5mm}

\noindent\textbf{Implicit Multi-modal Context Mining Submodule.}
The RGB and depth context mining submodules only consider a single-modal input. \jychecked{To model cross-modal contexts, we} propose the implicit multi-modal context mining submodule.
% , which aims to extract multi-modal rich contextual features implicitly by taking the fused multi-modal features as input. Specifically, this
% submodule takes the RGB backbone features $x^\text{RGB}$ and the depth backbone features $x^\text{depth}$ as input. These two input features are first concatenated and forwarded to a convolution layer to obtain the implicit multi-modal input features $x^\text{mul}$. 
\jychecked{Architecturally, this submodule has the same design as the RGB and the depth context mining submodules, \jychecked{and outputs} implicit multi-modal rich context features $\mathbb{ARC}^{\text{imp}}$.}
Like the first two submodules, we then apply channel-wise attention and context-wise attention on $\mathbb{ARC}^{\text{imp}}$ to obtain the final output $y^\text{imp}$ of this submodule.

\vspace{0.5mm}

\noindent\textbf{Explicit Multi-modal Context Mining Submodule.}
\jychecked{Since the implicit multi-modal context mining submodule only takes in the fused single-scale RGB and depth features and cannot disentangle multi-modal contexts in multiple scales, it is insufficient for modeling the contextual associations between different modalities.}
To capture the multi-modal contextual information in multiple scales explicitly, we propose the explicit multi-modal context mining submodule
% Unlike the implicit one, which takes the directly fused backbone features from the RGB and depth backbone networks as input, this submodule utilizes 
It utilizes the rich context features $\mathbb{RC}_{r_i,r_j}^{\text{RGB}}$ and $\mathbb{RC}_{r_i,r_j}^{\text{depth}}$ generated by the single-modal context mining submodules (\ie, RGB and depth context mining submodules). Each set of rich context features from the same scale are forwarded to a $3\times 3$ convolution with a $r_j$ dilation rate. The aggregated rich contextual features for the explicit multi-modal rich context features $\mathbb{ARC}^{\text{exp}}$ are computed as:
\begin{equation}
\begin{split}
        \mathbb{RC}_{r_i,r_j}^{\text{exp}}  &= \psi_{r_j}(\mathbb{RC}_{r_i, r_j}^{\text{RGB}} + \mathbb{RC}_{r_i, r_j}^{\text{depth}}) \\
        \mathbb{ARC}^{\text{exp}} &= \text{Concat}(\underbrace{\mathbb{RC}_{1,2}^{\text{exp}}, \mathbb{RC}_{1,4}^{\text{exp}}, ...,\mathbb{RC}_{2,8}^{\text{exp}}, \mathbb{RC}_{4,8}^{\text{exp}}
        }_{\binom{4}{2}}) ,
\end{split}
\end{equation}
\jychecked{Finally, we forward $\mathbb{ARC}^{\text{exp}}$ to our channel-wise attention and the context-wise attention to obtain the output $y^\text{exp}$.
}

\vspace{0.5mm}

\noindent\textbf{Adaptive Selection.}
\jychecked{Simply combining single-modal contextual features and multi-modal contextual features will cause a performance drop owing to the presence of a domain gap between different modalities. 
% For example, it would be challenging to predict glass surfaces from insufficient visual information (\eg, lack of context and weak reflection), while the depth information \jychecked{may be able to supplement this limitation}.
% In addition, rigidly selecting a particular set of contextual information from different modalities can reduce the generality of the proposed model, \jychecked{as the contextual information from different modalities may different cases in predicting the glass surfaces}.
}
\jychecked{Thus, we adaptively select the features from RGB context, depth context, implicit multi-modal context, and explicit multi-modal context information by dynamically adjusting the importance of different contextual features.}
To achieve this goal, we concatenate the outputs of these four context mining submodules as the input of our adaptive selection process, denoted as $x^\text{sel} \in \mathbb{R}^{4C \times H\times W}$. We feed $x^\text{sel}$ to a $1\times 1$ convolution layer to reduce its channel size to $C$. After that, we apply the channel-wise attention mechanism to these features in order to capture the importance of each channel. Based on the extracted channel-wise attention \jychecked{weights}, we can adaptively select the context features by multiplying \jychecked{these weights} to the input features to obtain the final cross-modal rich contextual features $\mathbb{CRC}$ as the output of the CCM module.

\subsection{Depth-missing Aware Attention (DAA) Module}\label{sec:dmaa}
\begin{figure}[t]
    \centering
    \includegraphics[width=0.45\textwidth]{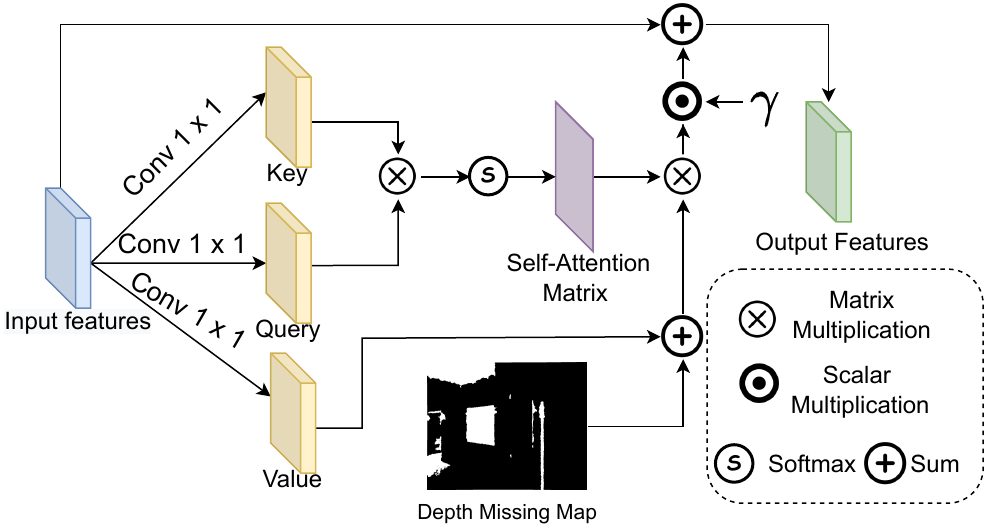}
    \vspace{-4mm}
    \caption{Illustration of our proposed \jychecked{DAA module}. We use three individual 1$\times$1 convolutions to generate key, query and value features from the input features. The key and query are multiplied and forwarded to Softmax to form the self-attention matrix. 
    The depth missing map is added to the value features, multiplied by the self-attention matrix and a learnable weight $\gamma$ to obtain the enhanced features, which are further added to input features to produce DAA's output.}
    \label{fig:dmaa}
    \vspace{-2mm}
\end{figure}

We observe that missing depth often appears around glass surfaces in the depth map \jychecked{due to light transmission, refraction and possibly reflection of the glass surface}. To exploit this cue, we propose a novel DAA module to explicitly involve the spatial information of the depth-missing regions into our framework. 
% The proposed DAA module takes the final output of the  CCM module $\mathbb{CRC}$, the output of the RGB context mining submodule $y^\text{RGB}$ and the output of the depth context mining submodule $y^\text{depth}$ in the CCM module as the input features. 
The proposed DAA module takes $\mathbb{CRC}$, $y^\text{RGB}$, and $y^\text{depth}$ as the input features. 
Besides, we also take the resized depth-missing map $Dm$
% of the same spatial resolution
as input. \fref{fig:dmaa} illustrates the structure of the proposed \jychecked{DAA module.}
Formally, the \jychecked{DAA module} is defined as:
\begin{equation}
\begin{split}
    f_k^{i} &= \varphi_v^{i}(x_{i}) + Dm, \\
    f_{out}^{i} &= \gamma^{i} \sigma(\varphi_q^{i}(x_{i})\varphi_k^{i}(x_{i})^{T}) f_k^{i} + x_{i},
\end{split}
\end{equation}
where $\varphi_q^{i}$, $\varphi_k^{i}$ and $\varphi_v^{i}$ are $1\times 1$ convolution layers for the $i$ modality features. $\sigma$ is a softmax function. $\gamma^{i}$ is a learnable weighting parameter for the $i$ modality features. $f_{out}^{i}$ are the output features for modality $i$, where $i \in \{\text{cm}, \text{RGB}, \text{depth}\}$ is the cross-modal, RGB and depth modality, respectively. 

\subsection{Loss Functions}\label{sec:loss}
We use a hybrid loss function $\mathcal{L} =  \sum_{i=1}^{N}{(\mathcal{L}_{BCE} + \mathcal{L}_{IoU})}$,
% , which combines the binary cross-entropy (BCE) loss and the interaction-over-union (IoU) loss to supervise the training of the multi-scale glass surface maps. The final loss function is:
% \begin{equation}
% Loss =  \sum_{i=1}^{N}{(L_{BCE} + L_{IoU})},
% \end{equation}
where $L_{BCE}$ and $L_{IoU}$ are the binary cross-entropy loss and the interaction-over-union loss, respectively, between the predicted
glass surfaces on the $i$-th stage and the ground truth glass surface map. $N$ is the number of stages.

\section{Experiments}

\begin{table*}[t]
    \centering
    \resizebox{.9\textwidth}{!}{
    \begin{tabular}{lc|cccc|cccc|cccc}
    \toprule
    \multirow{2}{*}{Methods} & \multirow{2}{*}{Venue} & \multicolumn{4}{c|}{RGB-D GSD} & \multicolumn{4}{c|}{GDD} & \multicolumn{4}{c}{GSD} \\ \cmidrule{3-14}  
                             & & IoU$\uparrow$ & F$_\beta\uparrow$ & MAE$\downarrow$ & BER$\downarrow$ & IoU$\uparrow$ & F$_\beta\uparrow$ & MAE$\downarrow$ & BER$\downarrow$ & IoU$\uparrow$ & F$_\beta\uparrow$ & MAE$\downarrow$ & BER$\downarrow$ \\ \midrule
    SPNet     &   \small{\textit{ICCV 2021}}    & 0.706    & 0.831    & 0.050   & 11.41  & -        & -               & -           & -           & -         & -               & -           & -            \\ 
    CLNet     &   \small{\textit{ICCV 2021}}    & 0.707    & 0.829    & 0.051   & 11.00  & -        & -               & -           & -           & -         & -               & -           & -            \\ 
    
    \midrule
    GDNet     &   \small{\textit{CVPR 2020}}    & 0.468    & 0.631    & 0.119   & 19.25  & 0.814         & 0.909               & 0.097           & 8.83            & 0.790         & 0.869               & 0.069           & 7.72            \\ 
    EBLNet    &   \small{\textit{ICCV 2021}}    & 0.707    & 0.819    & 0.048   & 10.91  & 0.870         & 0.922               & 0.064           & 6.08            & 0.817         & 0.878               & 0.059           & 6.75            \\ 
    GSDNet    &   \small{\textit{CVPR 2021}}    & 0.714    & 0.822    & 0.048   & 9.73   & 0.881         & 0.932               & \textbf{0.059}   & 5.71            & 0.836         & 0.903               & 0.055           & 6.12            \\ 
    RFENet    &   \small{\textit{IJCAI 2023}}   & 0.699    & 0.825    & 0.046   & 11.42  & 0.874         & 0.929               & 0.062           & 5.79            & 0.836         & 0.904               & \textbf{0.049}   & 6.24            \\ \midrule
    Ours      &                & \textbf{0.742} & \textbf{0.853} & \textbf{0.043} & \textbf{9.33} & \textbf{0.883} & \textbf{0.933} & \textbf{0.059} & \textbf{5.65} & \textbf{0.849} & \textbf{0.912} & \textbf{0.050} & \textbf{6.02} \\ 
    \bottomrule
    \end{tabular}
    }
\vspace{-3mm}
    \caption{Quantitative results on the RGB-D GSD, GDD, and GSD datasets.}
    \label{tab:combined_results}
\end{table*}

\noindent\textbf{\jychecked{Datasets} and Evaluation Metrics.}
\jychecked{We evaluate our method in two sets of experiments: one using our RGB-D dataset and the other using existing RGB glass datasets such as GDD and GSD. 
% The first set of experiments is to evaluate how well our proposed method performs when trained/tested under our proposed setting, \ie, based on RGB-D data, while the second set evaluates it under the existing setting, \ie, based on RGB data. 
% For our dataset, we use 2,400 images for training and 609 images for testing.
% We follow the training and test splits of the used datasets.
}
We use four metrics for evaluation, including intersection-over-union (IoU), F-measure ($F_\beta$), mean absolute error (MAE), and balance error rate (BER).

\begin{figure*}[t] \centering
    \renewcommand{\tabcolsep}{1pt}
    \newcommand{\imgpath}{select_images/0images/}
    \newcommand{\depthpath}{select_images/depths/}
    \newcommand{\CLNetpath}{select_images/ICCV2021-cascaded_rgbd_sod_results/}
    \newcommand{\SPNetpath}{select_images/ICCV2021-SPNet/}
    \newcommand{\GDNetpath}{select_images/CVPR2020-GDNet_depth/}
    \newcommand{\linpath}{select_images/CVPR2021-Lin/}
    \newcommand{\eblnetpath}{select_images/ICCV2021-EBLNet/}
    \newcommand{\rfenetpath}{select_images/RFENet/}
    \newcommand{\gtpath}{select_images/masks/}
    \newcommand{\ourpath}{select_images/ours_best_130/}
    \newcommand{\imgwidth}{0.09\textwidth}
    \begin{center}
    \begin{tabular}{cccccccccccc}
    % \gdef\imgid{a56d12d5543e4e69a47208426d21bea6_1_1}
    % \includegraphics[width=\imgwidth]{\imgpath \imgid .png} &
    % \includegraphics[width=\imgwidth]{\depthpath \imgid .png} &
    % \includegraphics[width=\imgwidth]{\CLNetpath \imgid .png} &
    % \includegraphics[width=\imgwidth]{\SPNetpath \imgid .png} &
    % \includegraphics[width=\imgwidth]{\GDNetpath \imgid .png} &
    % \includegraphics[width=\imgwidth]{\linpath \imgid .png} &
    % \includegraphics[width=\imgwidth]{\eblnetpath \imgid .png} &
    % \includegraphics[width=\imgwidth]{\ourpath \imgid .png} &
    % \includegraphics[width=\imgwidth]{\gtpath \imgid .png}
    % \\
    
    \gdef\imgid{00000642} %730x530
    \includegraphics[width=\imgwidth]{\imgpath \imgid .png} &
    \includegraphics[width=\imgwidth]{\depthpath \imgid .png} &
    \includegraphics[width=\imgwidth]{\CLNetpath \imgid .png} &
    \includegraphics[width=\imgwidth]{\SPNetpath \imgid .png} &
    \includegraphics[width=\imgwidth]{\GDNetpath \imgid .png} &
    \includegraphics[width=\imgwidth]{\linpath \imgid .png} &
    \includegraphics[width=\imgwidth]{\eblnetpath \imgid .png} &
    \includegraphics[width=\imgwidth]{\rfenetpath \imgid .png} &
    \includegraphics[width=\imgwidth]{\ourpath \imgid .png} &
    \includegraphics[width=\imgwidth]{\gtpath \imgid .png}
    \\

    \gdef\imgid{00003259}
    \includegraphics[width=\imgwidth]{\imgpath \imgid .png} &
    \includegraphics[width=\imgwidth]{\depthpath \imgid .png} &
    \includegraphics[width=\imgwidth]{\CLNetpath \imgid .png} &
    \includegraphics[width=\imgwidth]{\SPNetpath \imgid .png} &
    \includegraphics[width=\imgwidth]{\GDNetpath \imgid .png} &
    \includegraphics[width=\imgwidth]{\linpath \imgid .png} &
    \includegraphics[width=\imgwidth]{\eblnetpath \imgid .png} &
    \includegraphics[width=\imgwidth]{\rfenetpath \imgid .png} &
    \includegraphics[width=\imgwidth]{\ourpath \imgid .png} &
    \includegraphics[width=\imgwidth]{\gtpath \imgid .png}
    \\

    % \gdef\imgid{00004958}
    % \includegraphics[width=\imgwidth]{\imgpath \imgid .png} &
    % \includegraphics[width=\imgwidth]{\depthpath \imgid .png} &
    % \includegraphics[width=\imgwidth]{\CLNetpath \imgid .png} &
    % \includegraphics[width=\imgwidth]{\SPNetpath \imgid .png} &
    % \includegraphics[width=\imgwidth]{\GDNetpath \imgid .png} &
    % \includegraphics[width=\imgwidth]{\linpath \imgid .png} &
    % \includegraphics[width=\imgwidth]{\eblnetpath \imgid .png} &
    % \includegraphics[width=\imgwidth]{\ourpath \imgid .png} &
    % \includegraphics[width=\imgwidth]{\gtpath \imgid .png}
    % \\

    \gdef\imgid{9d26c238860b4508aaff3299af4c1681_2_1}
    \includegraphics[width=\imgwidth]{\imgpath \imgid .jpg} &
    \includegraphics[width=\imgwidth]{\depthpath \imgid .png} &
    \includegraphics[width=\imgwidth]{\CLNetpath \imgid .png} &
    \includegraphics[width=\imgwidth]{\SPNetpath \imgid .png} &
    \includegraphics[width=\imgwidth]{\GDNetpath \imgid .png} &
    \includegraphics[width=\imgwidth]{\linpath \imgid .png} &
    \includegraphics[width=\imgwidth]{\eblnetpath \imgid .png} &
    \includegraphics[width=\imgwidth]{\rfenetpath \imgid .png} &
    \includegraphics[width=\imgwidth]{\ourpath \imgid .png} &
    \includegraphics[width=\imgwidth]{\gtpath \imgid .png}
    \\

    % \gdef\imgid{08b7967373ca471e8b00cc96a3e8e2c1_1_2}
    % \includegraphics[width=\imgwidth]{\imgpath \imgid .png} &
    % \includegraphics[width=\imgwidth]{\depthpath \imgid .png} &
    % \includegraphics[width=\imgwidth]{\CLNetpath \imgid .png} &
    % \includegraphics[width=\imgwidth]{\SPNetpath \imgid .png} &
    % \includegraphics[width=\imgwidth]{\GDNetpath \imgid .png} &
    % \includegraphics[width=\imgwidth]{\linpath \imgid .png} &
    % \includegraphics[width=\imgwidth]{\eblnetpath \imgid .png} &
    % \includegraphics[width=\imgwidth]{\ourpath \imgid .png} &
    % \includegraphics[width=\imgwidth]{\gtpath \imgid .png}
    % \\  
    
    \gdef\imgid{camera_e4f9e0687dfa4a65892268c434e59720_lobby_1_frame_27_domain}
    \includegraphics[width=\imgwidth]{\imgpath \imgid .jpg} &
    \includegraphics[width=\imgwidth]{\depthpath \imgid .png} &
    \includegraphics[width=\imgwidth]{\CLNetpath \imgid .png} &
    \includegraphics[width=\imgwidth]{\SPNetpath \imgid .png} &
    \includegraphics[width=\imgwidth]{\GDNetpath \imgid .png} &
    \includegraphics[width=\imgwidth]{\linpath \imgid .png} &
    \includegraphics[width=\imgwidth]{\eblnetpath \imgid .png} &
    \includegraphics[width=\imgwidth]{\rfenetpath \imgid .png} &
    \includegraphics[width=\imgwidth]{\ourpath \imgid .png} &
    \includegraphics[width=\imgwidth]{\gtpath \imgid .png}
    \\

    \makebox[\imgwidth]{\footnotesize Image} &
    \makebox[\imgwidth]{\footnotesize Depth} &
    \makebox[\imgwidth]{\footnotesize CLNet} &
    \makebox[\imgwidth]{\footnotesize SPNet} &
    \makebox[\imgwidth]{\footnotesize GDNet} &
    \makebox[\imgwidth]{\footnotesize GSDNet} &
    \makebox[\imgwidth]{\footnotesize EBLNet} &
    \makebox[\imgwidth]{\footnotesize RFENet} &
    \makebox[\imgwidth]{\footnotesize Ours} &
    \makebox[\imgwidth]{\footnotesize GT}
    \\
    \end{tabular}
    \end{center}
    \vspace{-4mm}
    \caption{Visual comparison of our method with state-of-the-art methods on images \jychecked{from our RGB-D GSD dataset}. 
    % CLNet~\cite{cascaded_rgbd_sod} and SPNet~\cite{zhouiccv2021} are RGB-D salient object detection methods, while GDNet~\cite{Mei_2020_CVPR}, GSDNet~\cite{GSD:2021}, and EBLNet~\cite{He_2021_ICCV} are RGB-based glass surface detection methods.
    }
    \label{fig:visual}
    \vspace{-4mm}
    
\end{figure*}

\vspace{0.5mm}

\noindent\textbf{Implementation Details}
% Our proposed network is implemented using Pytorch. 
We use  ResNext-101~\cite{xie2017aggregated} pretrained on ImageNet as our backbone for the RGB image. 
% The details of our depth backbone network are shown in Table~\ref{table:depth_backbone}. 
We use the Adam optimzier~\cite{kingma:adam} with an initial learning rate of $1e-4$, training epoch 130 and batch size 14. 
% The initial learning rate is divided by 10 after 120 epochs.
% We resize all RGB images, depth maps and the corresponding ground truth masks to the spatial size of $400\times400$, and then randomly crop them to $384\times384$. To prevent overfitting, we adopt random horizontal flipping during our training process.
% We set the number of training epochs to 130, and the batch size used in training to 14. 
% We randomly initialize the parameters in all layers except the backbone network for RGB input images.
Note that we do not apply any post-processing technique (\eg, conditional random field (CRF)~\cite{krahenbuhl2011efficient_crf}) to our predicted maps for final output. 
Our model takes about 14 hours to converge, and 0.10s per image for inference on a single RTX2080Ti.

\vspace{0.5mm}

\noindent\textbf{Quantitative Evaluation.}
We evaluate the performance of the proposed method quantitatively on both RGB-D and RGB glass surface detection on the proposed RGB-D GSD dataset and the existing glass dataset GDD and GSD.
We compare the proposed method with SPNet~\cite{zhouiccv2021} and CLNet~\cite{cascaded_rgbd_sod} for RGB-D salient object detection; GDNet~\cite{Mei_2020_CVPR}, GSDNet\cite{GSD:2021}, EBLNet~\cite{He_2021_ICCV} and RFENet~\cite{RFENet} for \jychecked{RGB} glass surface detection. 
Additional results, including comparisons with more relevant baseline methods (\textit{e.g.}, MirrorNet~\cite{yang2019my} and PMD~\cite{lin2020progressive} for mirror detection), are provided in the Supplemental.
% and VGSD-Net~\cite{liu2024multi} 
For RGB-based methods, we use their publicly available codes with default configurations and we only use the RGB images and the ground truth masks for training and testing on our RGB-D GSD dataset. For RGB-D-based methods (\textit{e.g.}, SPNet and CLNet), we do not evaluate them on the RGB datasets (\textit{i.e.}, GDD and GSD) due to the absence of GT depth maps in these datasets.

To adopt our method to RGB glass surface detection, we keep the RGB backbone and the RGB context mining submodule used in the CCM module, with the depth backbone network and the DAA modules removed from our original framework.
\jychecked{Table~\ref{tab:combined_results} shows the experimental results.}
We can see that our proposed method significantly outperforms these baseline methods on all four metrics on our RGB-D GSD dataset.
In particular, our method shows substantial improvement on MAE, with a performance increase by 10.42\% over the second-best method GSDNet. 
\jychecked{Besides, our method outperforms all compared methods on both RGB glass surface detection datasets, GDD and GSD}, even though it does not use any auxiliary information as some glass surface detection methods do \jychecked{(\eg, \jychecked{boundary labels used in EBLNet}} and the reflection maps
used in GSDNet).
\jychecked{Specifically, our method achieves a significant performance improvement on the more challenging GSD dataset. The reason for our method to obtain a relatively minor performance gain on GDD dataset is that GDD contains images mostly captured from limited scenes and existing methods can perform very well on them. 
These results show that our method with RGB context mining submodules is particularly effective for glass surface detection in complex real-world scenes.}

\vspace{0.5mm}

\noindent\textbf{Qualitative Evaluation.} 
% We further demonstrate our method visually in \fref{fig:visual}, comparing with five state-of-the-art methods (due to space limitation, including two best-performing RGB-D salient object detection methods CLNet and SPNet, according to Table~\ref{tab:combined_results}, and all three existing glass surface detection methods).
We further demonstrate our method visually in \fref{fig:visual}, comparing it with four state-of-the-art glass surface detection methods.
The 1$st$ row of \fref{fig:visual} shows that state-of-the-art methods may struggle with \jychecked{visually ambiguous glass regions due to uncommon context}.
% \jychecked{Interestingly, both RGB-D methods, CLNet and SPNet, perform reasonably well, demonstrating the benefits of depth information for RGB-D glass surface detection.}
In these cases, using only RGB information to predict the glass surfaces can be challenging. 
In the 2$nd$ row, our method correctly detects over-exposed glass regions, 
\jychecked{by integrating both RGB and depth contextual information.}
% \jychecked{However, relying solely on RGB-D information without considering glass surface properties is unreliable, \jychecked{as shown by CLNet and SPNet}.}
% In the 3$rd$ row, both CLNet and SPNet only detect the depth-missing glass region, while our method accurately predicts the entire glass surface using cross-modal context mining.
The last row shows a challenging example with extremely tiny glass surface regions. Our method still significantly outperforms the baselines. \jychecked{We attribute it to the multi-scale context features incorporated in our proposed CCM module.}

\vspace{0.5mm}

\noindent\textbf{Ablation Study.}
Finally, we perform an ablation study to evaluate each of the proposed components of our model. Table~\ref{table:ablation} shows the results of our ablation study on our proposed dataset. We can see that adding either the CCM module or the DAA module \heichecked{(by replacing CMM with a single convolution layer to integrate multi-modal inputs)} helps improve the model performances, but our final model with both CCM and DAA modules performs the best on all four metrics. Note that adding the CCM modules to the base network (\ie, RGB and depth backbone networks with decoders) to form an ablated model (``B + CCM'') significantly outperforms the ablated model (``B + DAA'') that adds the DAA module to the base network. We attribute this to the success of our cross-modal contextual mining process conducted by the CCM module, which benefits the RGB-D glass surface detection task from a global view. 
Figure~\ref{fig:ablation} shows a visual example of the component analysis. We can see that the proposed CCM and DAA modules can help improve performance by exploiting cross-modal and depth-missing information to remove the over-predicted regions.
To verify the benefits of the depth cue in RGB-D glass surface detection, we conduct the following experiments: 1) remove all depth-related modules (\ie, only keep the RGB backbone and the RGB context mining submodule in CCM) and then retrain the proposed network 
\jychecked{(``Ours w/ RGB only'')}; 2) replacing the depth map with the grayscale image of the input RGB image during inference (\jychecked{``Ours} w/ RGB + Gray''); 3) replacing the depth map with a black \heichecked{(empty)} image during inference (\jychecked{``Ours} w/ RGB + Black''). 
% We can see that using only RGB information and removing depth-related modules (``Ours w/ RGB only'') produce unsatisfying results on our RGB-D GSD dataset.
We can see that \heichecked{removing depth-related modules and using only RGB information} 1) produces unsatisfying results on our RGB-D GSD dataset.
In addition, replacing the depth map with the grayscale version of the input image or with an empty map significantly decreases the model performance. 
These experimental results show the effectiveness and importance of the depth cue for RGB-D glass surface detection.

\begin{table}[t]
    \centering
    \resizebox{.4\textwidth}{!}{
    \begin{tabular}{@{}lcccc@{}}
    \toprule
        Methods & IoU$\uparrow$ & $F_\beta \uparrow$ & MAE$\downarrow$ & BER$\downarrow$ \\ 
    \midrule
        B &  0.703   &   0.814      &  0.046   &   11.14    \\
        B + CCM &  0.727   &   0.836      &  0.045   &   9.59    \\
        B + DAA &  0.706   &   0.819      &  0.046   &  10.48    \\ \midrule
        Ours w/ RGB only &  0.686   &   0.802      &  0.052   &   11.47    \\
        Ours w/ RGB + Gray &   0.325   &   0.646      &  0.112   &   33.02   \\
        Ours w/ RGB + Black &   0.352   &   0.639      &  0.112   &   32.87    \\\midrule
        Ours &  \textbf{0.742}   &   \textbf{0.853}      &  \textbf{0.043}   &   \textbf{9.33}\\
    \bottomrule
    \end{tabular}
    }
    \vspace{-3mm}
    \caption{Ablation studies. ``B'' is the base model. The upper section shows the ablation study on the effectiveness of CCM and DAA modules, while the lower section shows the study on the benefits of the depth cue. 
    % Best results are shown in bold.
    }
    \label{table:ablation}
\vspace{-4mm}
    
\end{table}

\begin{figure}[b]
\vspace{-3mm}
    \centering
    \renewcommand{\tabcolsep}{1pt}
    \newcommand{\imgpath}{visual_ablation/image/}
    \newcommand{\depthpath}{visual_ablation/depth/}
    \newcommand{\basepath}{visual_ablation/base/}
    \newcommand{\ccmpath}{visual_ablation/ccm/}
    \newcommand{\dmpath}{visual_ablation/dm/}
    \newcommand{\gtpath}{visual_ablation/gt/}
    \newcommand{\ourpath}{visual_ablation/ours/}
    \newcommand{\imgwidth}{0.065\textwidth}
    \begin{center}
    \begin{tabular}{@{}ccccccc@{}}
    \gdef\imgid{55bb9a9d764e41b98f7f1c1843885baa_0_5}
    \includegraphics[width=\imgwidth]{\imgpath \imgid .jpg} &
    \includegraphics[width=\imgwidth]{\depthpath \imgid .png} &
    \includegraphics[width=\imgwidth]{\basepath \imgid .png} &
    \includegraphics[width=\imgwidth]{\ccmpath \imgid .png} &
    \includegraphics[width=\imgwidth]{\dmpath \imgid .png} &
    \includegraphics[width=\imgwidth]{\ourpath \imgid .png} &
    \includegraphics[width=\imgwidth]{\gtpath \imgid .png}
    \vspace{-1.2mm}
    \\ 

    % \vspace{-2mm}
    \makebox[\imgwidth]{\scriptsize Image} &
    \makebox[\imgwidth]{\scriptsize Depth} &
    \makebox[\imgwidth]{\scriptsize B} &
    \makebox[\imgwidth]{\scriptsize B+CCM} &
    \makebox[\imgwidth]{\scriptsize B+DAA} &
    \makebox[\imgwidth]{\scriptsize Ours} &
    \makebox[\imgwidth]{\scriptsize GT}
    \\
    \end{tabular}
    \end{center}
    \vspace{-5mm}
    \caption{A visual example of the ablation study. 
    % The Base (``B'') model over-predicts the glass regions. 
    % With the CCM module, the model can largely exploit cross-modal information to reduce the over-prediction. 
    % \jychecked{The DAA module can also help reduce the triangular glass-like region near the center. As the region does not contain any depth missing pixels, it is less likely to be a glass surface.}
    % The full model of our method (including both CCM and DAA modules) performs the best in this example.
    }
    \label{fig:ablation}
\end{figure}

\textbf{Effectiveness of the DAA Module.} 
Table~\ref{table:DA} shows the ablation study on our proposed DAA module. ``DAA w/o $Dm$'' refers to the \jychecked{DAA module without taking the depth-missing map} as input. 
\jychecked{We design three other ablated models: ``DAA on RGB'', ``DAA on Depth'', and ``DAA on CM'' as adopting the DAA module 
only on
the RGB, depth, and cross-modal contextual features from the CCM modules, to test the effectiveness of the DAA module 
for extracting features
in different modalities.
}
Our final model adopts the DAA module with all three modalities. Experimental results show that the depth-missing information plays a key role in the DAA module, and our DAA module can effectively enhance the feature representation from different modalities.

\begin{table}[t]
    \centering
    \resizebox{.35\textwidth}{!}{
    \begin{tabular}{@{}lccccc@{}}
    \toprule
        Methods  & IoU$\uparrow$ & $F_\beta \uparrow$ & MAE$\downarrow$ & BER$\downarrow$ \\ 
    \midrule
        DAA w/o $Dm$ &  0.733   &   0.835      &  0.045   &   9.92    \\
        DAA on RGB &  0.738   &   0.838      &  0.044   &  9.62 \\
        DAA on Depth &  0.729   &   0.831      &  0.048   &  9.85 \\
        DAA on CM &  0.739   &   0.846      &  0.045   &  \textbf{9.30} \\ \midrule
        Ours &  \textbf{0.742}   &   \textbf{0.853}      &  \textbf{0.043}   &   9.33\\
    \bottomrule
    \end{tabular}
    }
    \vspace{-3mm}
    \caption{Ablation study of the \jychecked{DAA module}.
    \vspace{-4mm}
\label{table:DA}
% ,
% on our RGB-D GSD dataset. ``DAA w/o $Dm$'' refers to the DAA module without using the depth missing map as input. ``DAA on RGB/Depth/CM'' refers to the DAA module \jychecked{applied}
% \jychecked{on the RGB/depth/cross-modal contextual features extracted by the preceding CCM modules in \textit{stage4} and \textit{stage3}.}
% Best results are shown in bold.
}
\end{table}

\begin{figure}[b] 
    \centering
    \renewcommand{\tabcolsep}{1pt}
    \newcommand{\imgpath}{limitation/images/}
    \newcommand{\depthpath}{limitation/depths/}
    \newcommand{\gtpath}{limitation/masks/}
    \newcommand{\ourpath}{limitation/ours/}
    \newcommand{\imgwidth}{0.1\textwidth}
    \begin{center}
    \begin{tabular}{@{}ccccccc@{}}
    % \gdef\imgid{00003949}
    % \includegraphics[width=\imgwidth]{\imgpath \imgid .jpg} &
    % \includegraphics[width=\imgwidth]{\depthpath \imgid .png} &
    % \includegraphics[width=\imgwidth]{\ourpath \imgid .png} &
    % \includegraphics[width=\imgwidth]{\gtpath \imgid .png}
    % \vspace{-1.5mm}
    % \\ 
    
    % \gdef\imgid{00003414}
    % \includegraphics[width=\imgwidth]{\imgpath \imgid .jpg} &
    % \includegraphics[width=\imgwidth]{\depthpath \imgid .png} &
    % \includegraphics[width=\imgwidth]{\ourpath \imgid .png} &
    % \includegraphics[width=\imgwidth]{\gtpath \imgid .png}
    % \\ 

    \heichecked{} % added failure cases; replaced old 00003414 with following 3:
    \gdef\imgid{00003660}
    \includegraphics[width=\imgwidth]{\imgpath \imgid .jpg} &
    \includegraphics[width=\imgwidth]{\depthpath \imgid .png} &
    \includegraphics[width=\imgwidth]{\ourpath \imgid .png} &
    \includegraphics[width=\imgwidth]{\gtpath \imgid .png}
    \\ 
    \gdef\imgid{44ee8fadac69459e90be482c6a747e24_2_0}
    \includegraphics[width=\imgwidth]{\imgpath \imgid .jpg} &
    \includegraphics[width=\imgwidth]{\depthpath \imgid .jpg} &
    \includegraphics[width=\imgwidth]{\ourpath \imgid .png} &
    \includegraphics[width=\imgwidth]{\gtpath \imgid .png}
    \\ 
    \gdef\imgid{00002761}
    \includegraphics[width=\imgwidth]{\imgpath \imgid .jpg} &
    \includegraphics[width=\imgwidth]{\depthpath \imgid .png} &
    \includegraphics[width=\imgwidth]{\ourpath \imgid .png} &
    \includegraphics[width=\imgwidth]{\gtpath \imgid .png}
    \\ 
    \vspace{-4mm}
    \makebox[\imgwidth]{\scriptsize Image} &
    \makebox[\imgwidth]{\scriptsize Depth} &
    \makebox[\imgwidth]{\scriptsize Ours} &
    \makebox[\imgwidth]{\scriptsize GT}
    \\
    \end{tabular}
    \end{center}
    %\vspace{-2mm}
    % \caption{Failure cases. Our method may fail when both RGB images and depth maps lack contextual cues.}
    \caption{Failure cases. \heichecked{Our method may fail with low light scene, perfect glass transmission or reflection.}}
    \label{fig:failure}
\end{figure}

\section{Conclusion}
In this paper, we have investigated the glass surface detection problem by considering the depth information. To this end, we first construct a new large-scale RGB-D glass surface detection dataset containing 3,009 images with the corresponding depth maps and annotations. This dataset covers diverse scenes with glass surfaces and can facilitate research on glass surface detection. We then propose a novel RGB-D framework for glass surface detection. Our framework consists of two novel modules: (1) a cross-modal context mining (CCM) module for mining the context information among different modalities, and (2) a depth-missing aware attention (DAA) for exploiting the depth missing information around glass surfaces.
Experimental results show the superior performances of our proposed framework, compared with state-of-the-art methods from relevant fields.

% Despite the success, as our method focuses on mining context information in and across different modalities, \jychecked{it may fail if both RGB and depth information cannot provide sufficient contextual cues.
% Figure~\ref{fig:failure} shows that our method over-predicts glass-like regions (\ie, the blackboard) as glass surfaces, due to the lack of contextual cues in both RGB images and depth maps.} As future work, we are investigating to \jychecked{address the current failure cases by exploiting other cues for glass surface detection.} 
\heichecked{Despite its success, our method struggled under extreme conditions, such as low-light scenarios. In cases of perfect transmission, where depth signals fail to return and thus appear absent in depth scans, the depth context becomes irrelevant. Additionally, reflections from mirrors, which produce an almost identical image of the scene, continue to pose a significant challenge.}

% \clearpage

\section{Acknowledgments}
This work is in part supported by two GRF grants from the Research Grants Council of Hong Kong (RGC No.: 11211223 and 11220724).

\bibliography{aaai25}

% \clearpage
% \input{Reproducibility}

\end{document}

% ---------------------------------------------------------------
% TODO REVIEW: Replace with your title
\title{Supplementary Material: Leveraging RGB-D Data with Cross-Modal Context Mining for Glass Surface Detection} 

% TODO REVIEW: If the paper title is too long for the running head, you can set
% an abbreviated paper title here. If not, comment out.

\maketitle

\section{Overview}

This appendix is organized as follows:
\begin{itemize}
    \item We provide more details of our dataset.
    \item We provide more details of our method.
    \item More experimental results and ablation studies are included.
\end{itemize}

\section{RGB-D GSD Dataset}
\subsection{Dataset construction}
In Table~\ref{table:organization}, we show the composition of our proposed dataset.
We follow the dataset split of the original \ryn{datasets}.
\jychecked{For SUN RGB-D \cite{2017arXiv170201105A}, we have identified 630 images from the training set and 573 images from the test set of the original dataset, with glass surfaces. We then reallocate 290 images from the test set to the training set in order to keep the training-test ratio.}
For 2D-3D-Semantics \cite{7298655}, we follow cross validation fold \#2 (\ie, areas 1, 5, 6 for training and areas 2, 4 for testing). 
\jychecked{For Matterport3D \cite{Matterport3D}, we randomly split the selected images into a training set with 992 images and a test set with 214 images.} Refer to Table \ref{table:organization} for a summary of the composition of our dataset. 
Each RGB image is accompanied with a pre-processed depth image and finely annotated ground truth mask. The depth images were taken by different RGB-D cameras models, \eg, Asus Xtion, Kinect v2 \cite{2017arXiv170201105A} and Matterport \cite{Matterport3D} cameras. Although all depth images were encoded in 16-bit grayscale format, the definitions for missing depth \jychecked{are not the same in these three original datasets.} 
For example, in SUN RGB-D \cite{2017arXiv170201105A}, \jychecked{un-returned depth signals} were set to be the minimum value, which depends on the \jychecked{the depth ranges of individual images.}
On the other hand, 2D-3D-Semantics \cite{7298655} assumed invalid depth signals to be the maximum depth value (\ie, $2^{16} - 1$). We reorganize them and reset them to 0.

\begin{table}[htbp]
    \caption{Composition of our proposed RGB-D GSD dataset. We collect glass images from three existing RGB-D datasets. \jychecked{Note that as these datasets were originally created for other tasks, they do not include \jychecked{accurate} annotations of glass surface masks. Thus, we annotate the GT masks of the glass surfaces in our dataset construction.}}
    \label{table:organization}
    \centering
    \begin{tabular}{@{}cccc@{}}
    \toprule
        Dataset  & Whole & Train & Test\\ 
    \midrule
        SUN RGB-D & 1,203 & 920 & 283 \\
        2D-3D-Semantics    & 600   & 488 & 112  \\
        Matterport3D   & 1,206 & 992 & 214 \\
    \midrule
        Total & 3,009 & 2,400 & 609 \\
    \bottomrule
    \end{tabular}
\end{table}

% figure2
% \begin{figure*}
%   \centering \includegraphics[width=\textwidth]{charts/cropped_notused_demo_100_.png}
%   \caption{Examples from our RGB-D GSD dataset. Top, middle and bottom rows show RGB images, depth maps, and GT glass surface masks overlaid on the RGB images, respectively.}
%   \label{fig:glassrgbd}
% \end{figure*}

\section{Method}
As shown in Figure 4 in our main paper, the proposed framework consists of four major components: the backbone network for the input RGB images (in red), the backbone network for the input depth maps (in yellow), the cross-modal context mining (CCM) modules (in blue), and the depth-missing aware attention (DAA) modules (in green). These components are arranged to enable multi-stage feature learning with bottom-up and top-down information flows.

\textbf{Lighter Depth backbone.}The depth backbone network is much simpler and lighter, compared to the RGB one. There are two reasons.
First, using a lighter depth backbone network makes our full framework more efficient in both training and test stages. Second, we observe that depth maps contain sparser information. Simply adopting the same network as the RGB image for the depth map may cause a modality gap between the RGB and depth information, which will lead to performance degradation. Refer to Table~\ref{table:depth_backbone} for the detailed network architecture.

\begin{table}[t]
    \centering
    \caption{The architecture of the depth backbone network that we use for the input depth map. It consists of five stages, and each stage contains a convolution layer followed by a pooling layer. 
    Note that each ``conv-BR'' corresponds a sequence of convolution layer, BatchNorm layer and ReLU activation.  $K$, $S$ and $P$ denote the number of kernels, the number of strides and the padding size, respectively, used in the convolution layer.}
    \label{table:depth_backbone}
    \resizebox{.48\textwidth}{!}{
    \begin{tabular}{@{}ccc@{}}
    \toprule
        \thead{Layers Name}  & Layer Details & \thead{Output Size}\\ 
    \midrule
        Convolution & $3\times3$ conv-BR, $K=8$, $S=1$, $P=1$ & $384\times384$  \\
        Pooling & $2\times2$ max pool, stride 2 & $192\times192$  \\ \midrule
        Convolution & $3\times3$ conv-BR, $K=16$, $S=1$, $P=1$ & $192\times192$  \\
        Pooling & $2\times2$ max pool, stride 2 & $96\times96$  \\ \midrule
        Convolution & $3\times3$ conv-BR, $K=32$, $S=1$, $P=1$ & $96\times96$  \\
        Pooling & $2\times2$ max pool, stride 2 & $48\times48$  \\ \midrule
        Convolution & $3\times3$ conv-BR, $K=64$, $S=1$, $P=1$ & $48\times48$  \\
        Pooling & $2\times2$ max pool, stride 2 & $24\times24$  \\ \midrule
        Convolution & $3\times3$ conv-BR, $K=128$, $S=1$, $P=1$ & $24\times24$  \\
        Pooling & $2\times2$ max pool, stride 2 & $12\times12$  \\
    \bottomrule
    \end{tabular}
    }
\end{table}

\textbf{RGB Context Mining Submodule.}
The CNA mechanism that we use in the submodule consists of an average pooling layer and two convolution layers with a ReLU and sigmoid activation, as:
\begin{equation}
    CNA(x) = x \times \sigma(\psi_2(ReLU(\psi_1(\mu(x))))) ,
\end{equation}
\jychecked{where $\mu$, $ReLU$, $\sigma$ and $\psi$ are the global average pooling (GAP) layer, ReLU, sigmoid function and convolution layers with a $1\times 1$ kernel, respectively. $\psi_1$ and $\psi_2$ are $1\times 1$ convolution layers with different weights.}
$x$ represents the input features. The output of channel-wise attention has the same number of channels as the input features $x$.
Similarly, we can obtain the CXA by adjusting the output channels of the convolution layers that we use.

\textbf{Depth Context Mining Submodule.}
The difference between the RGB context mining submodule and the depth context mining submodule is that the former submodule takes the RGB backbone features $x^\text{RGB}$ as input, while the later submodule takes the depth backbone features $x^\text{depth}$. The model weights of these two submodules are not shared, so that they can focus on context mining in \jychecked{their own} modalities.

\textbf{Implicit Multi-modal Context Mining Submodule.}
The implicit multi-modal context mining submodule aims to extract multi-modal rich contextual features implicitly by taking the fused multi-modal features as input. Specifically, this
submodule takes the RGB backbone features $x^\text{RGB}$ and the depth backbone features $x^\text{depth}$ as input. These two input features are first concatenated and forwarded to a convolution layer to obtain the implicit multi-modal input features $x^\text{mul}$.

\textbf{Adaptive Selection.} 
For example, it would be challenging to predict glass surfaces from insufficient visual information (\eg, lack of context and weak reflection), while the depth information \jychecked{may be able to supplement this limitation}.
In addition, rigidly selecting a particular set of contextual information from different modalities can reduce the generality of the proposed model, \jychecked{as the contextual information from different modalities may different cases in predicting the glass surfaces}.

\section{Experiments}
\subsection{\jychecked{Datasets} and Evaluation Metrics}
\jychecked{For the evaluation, we use four metrics to evaluate the performances of our methods: intersection over union (IoU), F-measure, mean absolute error (MAE), and balance error rate (BER).} 
MAE is formulated as:
\begin{equation}
MAE=\frac{1}{HW} \sum_{i=1}^H  \sum_{j=1}^W{|P(i,j)-G(i,j)|},
\end{equation}
where $P$ is the predicted mask, and $G$ is ground truth. $H$ and $W$ are the width and height of the input image.

F-measure is calculated by a weighted combination of Precision and Recall:
\begin{equation}
F_\beta=\frac{1+\beta^2 (Precision \times Recall)}{\beta^2 Precision + Recall},
\end{equation}
where $\beta^2$ is set to 0.3 as suggested in~\cite{achanta2009frequency}.

The IoU score is calculated as:
\begin{equation}
IoU=\frac{N_{tp}}{N_{tp} + N_{fp} + N_{fn}},
\end{equation}
where $N_{tp}$, $N_{fp}$ and $N_{fn}$ are the numbers of true positive, false positive and false negative pixels, respectively.

The BER score is a widely used metric in shadow detection to measure the binary prediction from a balance-aware prospective, and is formulated as:
\begin{equation}
BER=1-0.5\times (\frac{N_{tp}}{N_p}+\frac{N_{tn}}{N_n}),
\end{equation}
where $N_{tp}$, $N_{tn}$, $N_p$, $N_n$ are the numbers of true positive, true negative, glass and non-glass pixels, respectively.

\subsection{Implementation Details}
Our proposed network is implemented using Pytorch. 
The details of our depth backbone network are shown in Table~\ref{table:depth_backbone}. 
% The initial learning rate is divided by 10 after 120 epochs.
We resize all RGB images, depth maps and the corresponding ground truth masks to the spatial size of $400\times400$, and then randomly crop them to $384\times384$. To prevent overfitting, we adopt random horizontal flipping during our training process.
% We set the number of training epochs to 130, and the batch size used in training to 14. 
We randomly initialize the parameters in all layers except the backbone network for RGB input images.

\subsection{More Results}

\jychecked{We perform two sets of experiments to evaluate the performance of the proposed method quantitatively. In our first set of experiments, we focus on RGB-D glass surface detection on our proposed RGB-D GSD dataset. We compare the proposed method with \jychecked{16} state-of-the-art methods from relevant fields \jychecked{on our RGB-D GSD} dataset. These methods include  DANet~\cite{ECCV2020_DANet}, BBS-Net~\cite{fan2020bbsnet}, DCF~\cite{Ji_2021_DCF}, SPNet~\cite{zhouiccv2021}, CLNet~\cite{cascaded_rgbd_sod}, MAGNet~\cite{zhong2024magnet} and Dformer~\cite{yin2024dformer} for RGB-D salient object detection and RGB-D semantic segmentation; MINet~\cite{MINet-CVPR2020}, GateNet~\cite{GateNet_eccv20}, CSNet~\cite{21PAMI-Sal100K} and PGNet~\cite{xie2022pyramid}, ADMNet~\cite{zhou2024admnet}  for RGB salient object detection;
GDNet~\cite{Mei_2020_CVPR}, GSDNet\cite{GSD:2021}, EBLNet~\cite{He_2021_ICCV} and RFENet~\cite{RFENet} 
% and VGSD-Net~\cite{liu2024multi} 
for \jychecked{RGB} glass surface detection. 
For these baseline methods, we use their publicly available codes with default configurations. \jychecked{For the RGB-based methods, we only use the RGB images and the ground truth masks for training and testing.}
For video-based methods, we duplicate the input images to adapt to their input sources.
}
Table~\ref{tab:sota_results} shows the experimental results.
We can see that our proposed method significantly outperforms these baseline methods on all four metrics.
In particular, our method shows substantial improvement on MAE, with a performance increase by 10.42\% over the second-best method GSDNet. 

In our second set of experiments, we study how well our proposed method performs when trained and tested on the existing RGB glass surface detection datasets (\ie, GDD~\cite{Mei_2020_CVPR} and GSD~\cite{GSD:2021}), \jychecked{which do not contain depth information for training and evaluation}.
\jychecked{We compare our method with relevant RGB-based methods, including BASNet~\cite{qin2019basnet}, MINet~\cite{MINet-CVPR2020}, GateNet~\cite{GateNet_eccv20}, CSNet~\cite{21PAMI-Sal100K}, and PGNet~\cite{xie2022pyramid} for salient object detection; 
MirrorNet~\cite{yang2019my} and PMD~\cite{lin2020progressive} for mirror detection;
GDNet, EBLNet, and GSDNet for glass surface detection.}
We use their official codes with the default configurations for all these methods.
\jychecked{We train and test all methods on the training/test splits of the same dataset.}
To adopt our method to RGB glass surface detection, we keep the RGB backbone and the RGB context mining submodule used in the CCM module, with the depth backbone network and the DAA modules removed from our original framework.
\jychecked{Table~\ref{tab:rgb_glass} shows the experimental results.}
\jychecked{We can see that our method outperforms all compared methods on both RGB glass surface detection datasets, GDD and GSD}, even though it does not use any auxiliary information as some glass surface detection methods do \jychecked{(\eg, \jychecked{boundary labels used in EBLNet}} and the reflection maps
used in GSDNet).
\jychecked{Specifically, our method achieves a significant performance improvement on the more challenging GSD dataset. The reason for our method to obtain a relatively minor performance gain on GDD dataset is that GDD contains images mostly captured from limited scenes and existing methods can perform very well on them. 
These results show that our method with RGB context mining submodules is particularly effective for glass surface detection in complex real-world scenes.}

\begin{table}[t]
    \centering
    \resizebox{0.3\textwidth}{!}{
    \begin{tabular}{@{}lcccc@{}}
    \toprule
        Methods & IoU$\uparrow$ & $F_\beta \uparrow$ & MAE$\downarrow$ & BER$\downarrow$ \\ 
    \midrule
        DANet & 0.636    &   0.791      &  0.063   &   14.94  \\
        BBS-Net & 0.662    &   0.808      &  0.055   &   14.24  \\ 
        DCF & 0.655 & 0.803 & 0.058 & 14.12  \\ 
        SPNet & 0.706    &   0.831      &  0.050   &   11.41  \\
        CLNet & 0.707 & 0.829 & 0.051 & 11.00  \\ 
        MAGNet & 0.716 & 0.830 & 0.045 & 10.07  \\ 
        DFormer & 0.672 & 0.829 & 0.050 & 12.74 \\
    \midrule
        MINet & 0.653 & 0.802 & 0.055 & 14.10 \\
        GateNet & 0.668 & 0.816 & 0.053 & 12.86 \\
        CSNet & 0.472 & 0.659 & 0.108 & 22.56 \\
        PGNet & 0.638 & 0.789 & 0.067 & 13.86 \\
        ADMNet & 0.522 & 0.661 & 0.097 & 20.30 \\
    \midrule
        GDNet   & 0.468    &   0.631      &  0.119   &   19.25  \\
        GSDNet    & 0.714    &  0.822       & 0.048    &   9.73  \\ 
        EBLNet   & 0.707    &  0.819       & 0.048    &   10.91  \\ 
        RFENet & 0.699 & 0.825 & 0.046 & 11.42 \\
        % VGSD-Net & 0.403    &  0.519       & 0.203    &   20.27 \\
    \midrule
        Ours  &  \textbf{0.742}   &   \textbf{0.853}      &  \textbf{0.043}   &   \textbf{9.33}\\
    \bottomrule
    \end{tabular}
    }
\vspace{-3mm}
    
        \caption{Quantitative results on our RGB-D GSD dataset. }
    \label{tab:sota_results}
\end{table}

\begin{table}[t]
% \vspace{-4mm}
	\centering

	\resizebox{.48\textwidth}{!}{
	\begin{tabular}{lcccccccc}
	\toprule
	\multirow{2}{*}{Methods}                                                                                                           & \multicolumn{4}{c}{GDD}                                     & \multicolumn{4}{c}{GSD}                                         \\ \cmidrule{2-9} 
							 & IoU$\uparrow$ & F$_\beta\uparrow$ & MAE$\downarrow$ & BER$\downarrow$ & IoU$\uparrow$ & F$_\beta\uparrow$ & MAE$\downarrow$ & BER$\downarrow$ \\ \midrule
	BASNet      & 0.808         & 0.891               & 0.106           & 9.37            & 0.698         & 0.808               & 0.106           & 13.54           \\ 
	
	MINet         & 0.844         & 0.919               &    0.077        & 7.40            & 0.773         & 0.879               & 0.077           & 9.54           \\ 
	GateNet        & 0.817         & 0.931               & 0.073           & 8.84            & 0.689            & 0.898                  & 0.073              & 10.12 \\ 
	CSNet  &  0.773 & 0.876 & 0.135 & 11.33 & 0.666 & 0.805 & 0.135 & 14.76\\
	PGNet & 0.857 & 0.930 & 0.074 & 6.82 & 0.805 & 0.897 & 0.068 & 7.88\\ \midrule
	MirrorNet       & 0.851         & 0.903               & 0.083           & 7.67            & 0.742         & 0.828               & 0.090           & 10.76           \\ 
	PMD          & 0.870         & 0.930               & 0.067           & 6.17            & 0.817         & 0.890               & 0.061           & 6.74           \\ 
 \midrule
	GDNet                    & 0.814         & 0.909               & 0.097           & 8.83            & 0.790         & 0.869               & 0.069           & 7.72            \\ 
	EBLNet   & 0.870         & 0.922               & 0.064           & 6.08            & 0.817        & 0.878                  & 0.059              & 6.75 \\ 
	GSDNet               & 0.881         & 0.932               & \textbf{0.059}           & 5.71            & 0.836         & 0.903               & 0.055           & 6.12            \\ 
	RFENet               & 0.874         & 0.929               & 0.062           & 5.79            & 0.836         & 0.904               & \textbf{0.049}           & 6.24            \\ \midrule
	Ours       &    \textbf{0.883}  &   \textbf{0.933} &  \textbf{0.059} &    \textbf{5.65} & \textbf{0.849} &    \textbf{0.912} & \textbf{0.050} & \textbf{6.02}    \\ 
    \bottomrule
\end{tabular}
}

\vspace{-3mm}
	\caption{Quantitative results of our method with the state-of-the-art RGB-based models on two existing glass surface detection datasets, GDD and GSD.}
	\label{tab:rgb_glass}
\end{table}

\subsection{Ablation Study}
\textbf{Effectiveness of the CCM Module.} 
Table~\ref{table:CCM} shows the ablation study on the proposed CCM module. Specifically, we keep all other modules in the final model while replacing our proposed CCM module with its variants, where ``RGB'', ``D'', ``imp.'', and ``exp.'' refer to the RGB context mining submodule, depth context mining submodule, implicit multi-modal context mining submodule, and explicit multi-modal context mining submodule, respectively, in the CCM module. We can see that the single-modal variants (\ie, ``CCM w/ RGB'' and ``CCM w/ D'') have the worse performances, compared with the other three multi-modal variants. 
We also observe that the ablated models with the cross-modal context mining submodule (\ie, ``CCM w/ RGB + D + imp.'' and ``CCM w/RGB + D + exp.'') outperform those without the submodules (\eg, ``CCM w/RGB + D'').
This indicates the importance of \jychecked{cross-modal context modeling} in our CCM module. Finally, our final model performs the best among all ablated models, which shows that the CCM module with cross-modal mining can provide a great performance improvement in glass surface detection.

\begin{table}[t]
\caption{Ablation study of the CCM module, on our RGB-D GSD dataset. ``RGB'', ``D'', ``imp.'', and ``exp.'' refer to the RGB context mining submodule, depth context mining submodule, implicit multi-modal context mining submodule, and explicit multi-modal context mining submodule, respectively, in the CCM module. ``CCM w/ RGB'' refers to the CCM module \jychecked{containing only the RGB context mining submodule, while ``CCM w/ RGB + D'' refers to the CMM module with both RGB and depth context mining submodules.}
}
\label{table:CCM}
    \centering
    \begin{tabular}{@{}lccccc@{}}
    \toprule
        Methods  & IoU$\uparrow$ & $F_\beta \uparrow$ & MAE$\downarrow$ & BER$\downarrow$ \\ 
    \midrule
        CCM w/ RGB &  0.708   &   0.819      &  0.046   &   10.44 \\
        CCM w/ D &  0.695   &   0.815      &  0.053   &   10.92    \\
        CCM w/ RGB + D &  0.716   &   0.827      &  0.047   &   10.18    \\
        CCM w/ RGB + D + imp. &  0.736   &   0.839      &  0.046   &   9.66    \\
        CCM w/ RGB + D + exp. &  0.737   &   0.841      &  0.043   &   9.65    \\ \midrule
        Ours &  \textbf{0.742}   &   \textbf{0.853}      &  \textbf{0.043}   &   \textbf{9.33}\\
    \bottomrule
\end{tabular}
\end{table}

\textbf{Effectiveness of the DAA Module.} 
Table~\ref{table:DA} shows the ablation study on our proposed DAA module. ``DAA w/o $Dm$'' refers to the \jychecked{DAA module without taking the depth-missing map} as input. 
\jychecked{We design three other ablated models: ``DAA on RGB'', ``DAA on Depth'', and ``DAA on CM'' as adopting the DAA module 
only on
the RGB, depth, and cross-modal contextual features from the CCM modules, to test the effectiveness of the DAA module 
for extracting features
in different modalities.
}
Our final model adopts the DAA module with all three modalities. Experimental results show that the depth-missing information plays a key role in the DAA module, and our DAA module can effectively enhance the feature representation from different modalities.

\begin{table}[t]
\caption{Ablation study of the \jychecked{DAA module},
on our RGB-D GSD dataset. ``DAA w/o $Dm$'' refers to the DAA module without using the depth missing map as input. ``DAA on RGB/Depth/CM'' refers to the DAA module \jychecked{applied}
\jychecked{on the RGB/depth/cross-modal contextual features extracted by the preceding CCM modules in \textit{stage4} and \textit{stage3}.}
Best results are shown in bold.}
\label{table:DA}
    \centering
    \begin{tabular}{@{}lccccc@{}}
    \toprule
        Methods  & IoU$\uparrow$ & $F_\beta \uparrow$ & MAE$\downarrow$ & BER$\downarrow$ \\ 
    \midrule
        DAA w/o $Dm$ &  0.733   &   0.835      &  0.045   &   9.92    \\
        DAA on RGB &  0.738   &   0.838      &  0.044   &  9.62 \\
        DAA on Depth &  0.729   &   0.831      &  0.048   &  9.85 \\
        DAA on CM &  0.739   &   0.846      &  0.045   &  \textbf{9.30} \\ \midrule
        Ours &  \textbf{0.742}   &   \textbf{0.853}      &  \textbf{0.043}   &   9.33\\
    \bottomrule
\end{tabular}
\end{table}

\subsection{Qualitative Evaluation}
We further demonstrate our method visually in \ref{fig:visual}, comparing it with five state-of-the-art methods.

\begin{figure*}[ht] \centering
    \renewcommand{\tabcolsep}{1pt}
    \newcommand{\imgpath}{select_images/0images/}
    \newcommand{\depthpath}{select_images/depths/}
    \newcommand{\CLNetpath}{select_images/ICCV2021-cascaded_rgbd_sod_results/}
    \newcommand{\SPNetpath}{select_images/ICCV2021-SPNet/}
    \newcommand{\GDNetpath}{select_images/CVPR2020-GDNet_depth/}
    \newcommand{\linpath}{select_images/CVPR2021-Lin/}
    \newcommand{\eblnetpath}{select_images/ICCV2021-EBLNet/}
    \newcommand{\gtpath}{select_images/masks/}
    \newcommand{\ourpath}{select_images/ours_best_130/}
    \newcommand{\imgwidth}{0.105\textwidth}
    \begin{center}
    \begin{tabular}{ccccccccc}
    \gdef\imgid{a56d12d5543e4e69a47208426d21bea6_1_1}
    \includegraphics[width=\imgwidth]{\imgpath \imgid .jpg} &
    \includegraphics[width=\imgwidth]{\depthpath \imgid .png} &
    \includegraphics[width=\imgwidth]{\CLNetpath \imgid .png} &
    \includegraphics[width=\imgwidth]{\SPNetpath \imgid .png} &
    \includegraphics[width=\imgwidth]{\GDNetpath \imgid .png} &
    \includegraphics[width=\imgwidth]{\linpath \imgid .png} &
    \includegraphics[width=\imgwidth]{\eblnetpath \imgid .png} &
    \includegraphics[width=\imgwidth]{\ourpath \imgid .png} &
    \includegraphics[width=\imgwidth]{\gtpath \imgid .png}
    \\
    
    \gdef\imgid{00000642}
    \includegraphics[width=\imgwidth]{\imgpath \imgid .png} &
    \includegraphics[width=\imgwidth]{\depthpath \imgid .png} &
    \includegraphics[width=\imgwidth]{\CLNetpath \imgid .png} &
    \includegraphics[width=\imgwidth]{\SPNetpath \imgid .png} &
    \includegraphics[width=\imgwidth]{\GDNetpath \imgid .png} &
    \includegraphics[width=\imgwidth]{\linpath \imgid .png} &
    \includegraphics[width=\imgwidth]{\eblnetpath \imgid .png} &
    \includegraphics[width=\imgwidth]{\ourpath \imgid .png} &
    \includegraphics[width=\imgwidth]{\gtpath \imgid .png}
    \\

    \gdef\imgid{00003259}
    \includegraphics[width=\imgwidth]{\imgpath \imgid .png} &
    \includegraphics[width=\imgwidth]{\depthpath \imgid .png} &
    \includegraphics[width=\imgwidth]{\CLNetpath \imgid .png} &
    \includegraphics[width=\imgwidth]{\SPNetpath \imgid .png} &
    \includegraphics[width=\imgwidth]{\GDNetpath \imgid .png} &
    \includegraphics[width=\imgwidth]{\linpath \imgid .png} &
    \includegraphics[width=\imgwidth]{\eblnetpath \imgid .png} &
    \includegraphics[width=\imgwidth]{\ourpath \imgid .png} &
    \includegraphics[width=\imgwidth]{\gtpath \imgid .png}
    \\

    \gdef\imgid{00004958}
    \includegraphics[width=\imgwidth]{\imgpath \imgid .png} &
    \includegraphics[width=\imgwidth]{\depthpath \imgid .png} &
    \includegraphics[width=\imgwidth]{\CLNetpath \imgid .png} &
    \includegraphics[width=\imgwidth]{\SPNetpath \imgid .png} &
    \includegraphics[width=\imgwidth]{\GDNetpath \imgid .png} &
    \includegraphics[width=\imgwidth]{\linpath \imgid .png} &
    \includegraphics[width=\imgwidth]{\eblnetpath \imgid .png} &
    \includegraphics[width=\imgwidth]{\ourpath \imgid .png} &
    \includegraphics[width=\imgwidth]{\gtpath \imgid .png}
    \\

    \gdef\imgid{9d26c238860b4508aaff3299af4c1681_2_1}
    \includegraphics[width=\imgwidth]{\imgpath \imgid .jpg} &
    \includegraphics[width=\imgwidth]{\depthpath \imgid .png} &
    \includegraphics[width=\imgwidth]{\CLNetpath \imgid .png} &
    \includegraphics[width=\imgwidth]{\SPNetpath \imgid .png} &
    \includegraphics[width=\imgwidth]{\GDNetpath \imgid .png} &
    \includegraphics[width=\imgwidth]{\linpath \imgid .png} &
    \includegraphics[width=\imgwidth]{\eblnetpath \imgid .png} &
    \includegraphics[width=\imgwidth]{\ourpath \imgid .png} &
    \includegraphics[width=\imgwidth]{\gtpath \imgid .png}
    \\

    \gdef\imgid{08b7967373ca471e8b00cc96a3e8e2c1_1_2}
    \includegraphics[width=\imgwidth]{\imgpath \imgid .jpg} &
    \includegraphics[width=\imgwidth]{\depthpath \imgid .png} &
    \includegraphics[width=\imgwidth]{\CLNetpath \imgid .png} &
    \includegraphics[width=\imgwidth]{\SPNetpath \imgid .png} &
    \includegraphics[width=\imgwidth]{\GDNetpath \imgid .png} &
    \includegraphics[width=\imgwidth]{\linpath \imgid .png} &
    \includegraphics[width=\imgwidth]{\eblnetpath \imgid .png} &
    \includegraphics[width=\imgwidth]{\ourpath \imgid .png} &
    \includegraphics[width=\imgwidth]{\gtpath \imgid .png}
    \\  
    
    \gdef\imgid{camera_e4f9e0687dfa4a65892268c434e59720_lobby_1_frame_27_domain}
    \includegraphics[width=\imgwidth]{\imgpath \imgid .jpg} &
    \includegraphics[width=\imgwidth]{\depthpath \imgid .png} &
    \includegraphics[width=\imgwidth]{\CLNetpath \imgid .png} &
    \includegraphics[width=\imgwidth]{\SPNetpath \imgid .png} &
    \includegraphics[width=\imgwidth]{\GDNetpath \imgid .png} &
    \includegraphics[width=\imgwidth]{\linpath \imgid .png} &
    \includegraphics[width=\imgwidth]{\eblnetpath \imgid .png} &
    \includegraphics[width=\imgwidth]{\ourpath \imgid .png} &
    \includegraphics[width=\imgwidth]{\gtpath \imgid .png}
    \\

    \makebox[\imgwidth]{\footnotesize Image} &
    \makebox[\imgwidth]{\footnotesize Depth} &
    \makebox[\imgwidth]{\footnotesize CLNet} &
    \makebox[\imgwidth]{\footnotesize SPNet} &
    \makebox[\imgwidth]{\footnotesize GDNet} &
    \makebox[\imgwidth]{\footnotesize GSDNet} &
    \makebox[\imgwidth]{\footnotesize EBLNet} &
    \makebox[\imgwidth]{\footnotesize Ours} &
    \makebox[\imgwidth]{\footnotesize GT}
    \\
    \end{tabular}
    \end{center}
    \caption{Visual comparison of our method with state-of-the-art methods on images \jychecked{from our RGB-D GSD dataset}. CLNet~\cite{cascaded_rgbd_sod} and SPNet~\cite{zhouiccv2021} are RGB-D salient object detection methods, while GDNet~\cite{Mei_2020_CVPR}, GSDNet~\cite{GSD:2021}, and EBLNet~\cite{He_2021_ICCV} are RGB-based glass surface detection methods.}
    \label{fig:visual}
\end{figure*}

\clearpage
\bibliography{aaai25}